\documentclass[10pt,twocolumn,letterpaper]{article}

\usepackage[pagenumbers]{cvpr} 
\usepackage{listings} 

\usepackage[dvipsnames]{xcolor}

\definecolor{cvprblue}{rgb}{0.21,0.49,0.74}

\usepackage{hyperref}
\hypersetup{pagebackref,breaklinks,colorlinks,citecolor=cvprblue}
\usepackage{microtype}

\def\paperID{368} 
\def\confName{3DV\xspace}
\def\confYear{2025\xspace}

\title{Dream-in-Style: Text-to-3D Generation Using Stylized Score Distillation}

\author{Hubert Kompanowski\\
Trinity College Dublin\\
\and
Binh-Son Hua\\
Trinity College Dublin\\
}

\begin{document}
\maketitle

\begin{abstract}
We present a method to generate 3D objects in styles. Our method takes a text prompt and a style reference image as input and reconstructs a neural radiance field to synthesize a 3D model with the content aligning with the text prompt and the style following the reference image. 
To simultaneously generate the 3D object and perform style transfer in one go, we propose a stylized score distillation loss to guide a text-to-3D optimization process to output visually plausible geometry and appearance. 
Our stylized score distillation is based on a combination of an original pretrained text-to-image model and its modified sibling with the key and value features of self-attention layers manipulated to inject styles from the reference image. 
Comparisons with state-of-the-art methods demonstrated the strong visual performance of our method, further supported by the quantitative results from our user study. 
\end{abstract}    
\section{Introduction}
\label{sec:intro}

Creating 3D content has been a key but demanding task in computer graphics. 
Traditional interactive tools such as Maya~\citep{maya}, Blender~\citep{blender} are among the most popular choices for novices and professionals to perform 3D modeling. 
In the wave of generative AI development, there have been increased interests in automatic synthesis of 3D content using generative models~\citep{poole2022dreamfusion,liu2023zero1to3}.  
This is an open research area with tremendous progress in recent years, with the rise of language models enabling the widespread adoption of natural languages to condition the automatic generation of data in different modalities.

This trend has stimulated the development of text-to-3D generation methods~\citep{poole2022dreamfusion,wang2023score}, where 3D objects can be generated by simply prompting an input sentence that describes the desired object content.
These methods are generic to the appearance of the generated objects, which means that the final look and feel of the 3D content is barely controllable. 
This is in contrast to a common requirement in traditional 3D modeling, where a visual artist might aim to decorate a 3D object in particular styles. 
For example, one might be interested in creating a 3D object with low polygon count, making its geometry appear as a collection of flat surfaces, or a 3D object with stylized textures in photorealistic or cartoon styles. 
Performing such a stylization using traditional tools is a tedious task. 
Therefore, integrating stylization into generative models is a promising idea to explore. 

At its core, text-to-3D generation~\citep{poole2022dreamfusion,wang2023prolificdreamer} performs an iterative update on a 3D representation such that its rendering converges to a photorealistic image scored by a pretrained text-to-image model. 
So far, most development of text-to-3D generation has focused on objects with generic appearance. 
Creating 3D content with particular styles remains challenging to achieve. 
A straightforward approach is to incorporate style description into text prompts used to generate 3D content, but this approach is not effective due to the ambiguity in how styles can be described using natural languages. 

\begin{figure*}[t]
    \centering
    \includegraphics[width=0.95\linewidth]{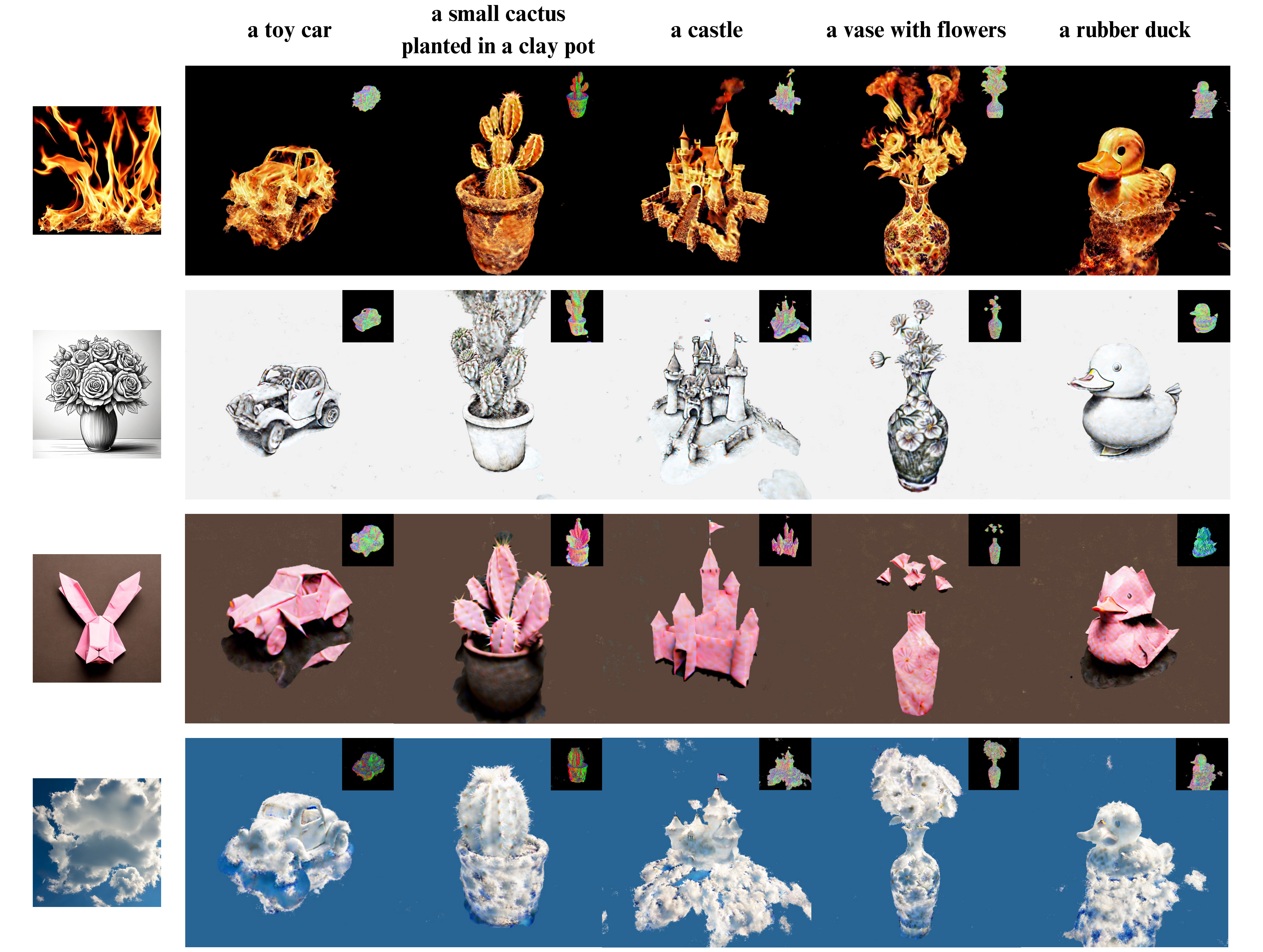}
    \caption{We aim to generate a 3D object jointly from a text prompt and a style reference image so that the object integrates the descriptive elements in the text prompt and the aesthetic style in the reference image. Our method employs a stylized score distillation to steer a text-to-3D optimization using combined scores from a pretrained text-to-image model and its modified variant on attention layer features, generating stylized 3D objects in a single-stage optimization.}
    \label{fig:teaser}
\end{figure*}

In this paper, we propose a new stylization method for 3D content creation from text prompts. 
Our method uses a style reference image to guide text-to-3D generation, transferring the detailed visual elements such as color, tone, or texture in the reference image to the final 3D object. 
This design choice is made to maximize style guidance so that the desired style can be described by both the reference image and the text prompt. 
Our method follows an optimization-based text-to-3D framework, performing a gradient descent update to optimize a 3D representation. 
Our update is regularized by a style-based score distillation that works as a critique to the rendered 3D content using a style-aware text-to-image model modified from the original pretrained model without any finetuning.
We formulate this process using a stylized score distillation gradient, which dynamically combines scores from both the original and modified pretrained model.  
Our experiments demonstrate the effectiveness of the proposed method along with its flexibility and robustness in the styles of the generated 3D content. 
Some example results are shown in \Cref{fig:teaser}.

In summary, our contributions are threefold. 
First, we propose to adapt a generic pretrained text-to-image model with a reference style image and build a training-free modified pretrained model for stylized text-to-3D generation. 
Second, we propose a stylized score distillation gradient to steer the 3D generation toward the desired style specified in the reference style image. 
Third, we demonstrate the flexibility of our method by applying our stylized distillation to different text-to-3D generation losses, including score distillation sampling (SDS)~\citep{poole2022dreamfusion}, noise-free score distillation~\citep{katzir2023noisefree}, and variational score distillation (VSD)~\citep{wang2023prolificdreamer}. We also demonstrate the robustness of our 3D stylization via a diverse set of text prompts and styles. 

\section{Related works}

\subsection{Style transfer}
Style transfer is a traditional computer graphics problem that aims to synthesize images in artistic styles. 
Example-based methods such as image analogies~\citep{hertzmann2001analogies,liao2017analogies} learn to filter a pair of source images so that when applied to a new target image, the filter can generate analogous filtered results. 
Image analogies are built upon texture synthesis, which requires source pairs to approximately align to learn effective filters. 
This restriction is alleviated in modern deep-learning based style transfer methods. 
Neural style transfer~\citep{gatys2016image} optimizes the target image so that it shares the content of an input image and the style of a reference image with feature correspondences characterized by a pretrained neural network. 
Several extensions of style transfer follow, notably for efficiency improvement with a feed-forward neural network~\citep{johnson2016perceptual} and image-to-image translation networks~\citep{isola2017image,zhu2017unpaired,liu2017unsupervised}, style representation with statistical features~\citep{huang2017arbitrary}, style representation using text-image features~\citep{gal2022stylegan,patashnik2021styleclip,kwon2022clipstyler}, and video style transfer~\citep{li2019linear}. 
We refer the reader to the comprehensive survey paper~\citep{jing2019neural} for a broader coverage on visual style transfer methods. 

\subsection{Generative content creation}
Generative models such as generative adversarial networks~\citep{goodfellow2014generative,karras2019style} and diffusion models~\citep{ho2020denoising,po2023survey} are notable tools to synthesize realistic data of different modalities, notably text and images when trained on large-scale datasets. 
Recently, text-to-image models like DALL-E~\citep{dalle3_2023} and Stable Diffusion~\citep{rombach2022high} have shown great promise in generating photorealistic images from arbitrary text prompts. 
Text-to-image diffusion models can be used for image editing~\citep{meng2021sdedit} including style transfer by learning a copy of a pretrained diffusion model~\citep{zhang2023adding}, bridging the latent space of two diffusion models for image-to-image translations~\citep{su2022dual,li2023bbdm}, and instruction-based translations supervised using image-prompt-image datasets~\citep{brooks2023instructpix2pix} and test-time editing directions~\citep{parmar2023zero}.
Personalized text-to-image methods~\citep{ruiz2022dreambooth} can generate images in similar subjects defined by a small set of reference images, but these methods requires multiple images to finetune the pretrained diffusion models. 
Textual inversion methods~\citep{gal2022textual,han2023hiper} instead only optimize the text embedding to obtain a text prompt that preserves the subject identity of an input image. 
Training-free methods~\citep{hertz2023stylealigned,jeong2024visual} inject reference features into the denoising process of a pretrained diffusion model through manipulating features input to attention layers of the denoising U-Net to influence the stylized generation.


In the 3D domain, generative models can be trained to sample 3D data represented by voxels~\citep{wu2015shape} and point clouds~\citep{nichol2022point}, but high-quality 3D training datasets are relatively scarce and are in smaller scales compared to language and image datasets~\citep{deitke2023objaverse,deitke2023objaverseXL}. 
A recent approach to sidestep this issue is to generate 3D data by learning from images. 
3D-aware GANs~\citep{chan2022efficient} learn to generate 3D-consistent images by incorporating a neural radiance field~\citep{mildenhall2020nerf} as the intermediate 3D representation, but their results are limited to a few categories of objects. 
Instead of focusing on 3D generation, novel view synthesis~\citep{chan2023genvs,szymanowicz23viewset} predicts 3D-consistent views from a sparse set of input views, with support from pretrained image diffusion models~\citep{liu2023zero1to3}.
Large reconstruction models~\citep{hong2024lrm,xu2024dmv3d} demonstrate effective image-to-3D generation by scaling up the training to millions of 3D objects. 
Text-to-3D generation~\citep{poole2022dreamfusion,wang2023score,wang2023prolificdreamer} is a recent advance aiming to generate 3D objects directly from text prompts by using a pretrained text-to-image model to score the rendering of 3D objects at random angles.
Text-to-3D methods have witnessed rapid development recently, with significant advances made toward improved distillation~\citep{katzir2023noisefree}, shape quality~\citep{lin2023magic3d,xu2023dream3d,chen2023fantasia3D,zhu2024hifa} with textures~\citep{metzer2023latent}, fast rendering~\citep{tang2024dreamgaussian}, amortized sampling~\citep{lorraine2023att3d,xie2024latte3d}, 3D editing~\citep{hertz2023delta,koo2024posterior}, and animated models~\citep{bahmani20244dfy}. 
Our method belongs to the text-to-3D family, but focuses on stylized 3D generation. 


\subsection{3D stylization}



Several approaches for 3D stylization exist. 
Traditionally, image analogies can be adapted to stylize 3D rendering while preserving physically based illumination effects~\citep{fiser2016stylit,sykora2019styleblit}.
Recent notable advances in neural radiance fields (NeRFs)~\citep{mildenhall2020nerf} allow us to perform stylization on implicit neural scene representations. 
NeRF stylization methods~\citep{nguyen2022snerf,fan2022unified,liu2023stylerf,zhang2022arf,pang2023locally} often assume a two-stage process in which a radiance field is first reconstructed on photorealistic images and then stylized based on style reference images. 
This process can be improved by several techniques including optimizing semantic correspondences between the radiance field and the style reference image~\citep{zhang2022arf,pang2023locally} or generalizing the stylization across scenes and styles with generalizable NeRFs~\citep{chiang2022stylizing,huang2022stylizedNeRF}. 
Two-stage stylization methods tend to change only the appearance of the NeRF while keeping the geometry intact~\citep{chiang2022stylizing,huang2022stylizedNeRF} because these methods lack a generative prior and therefore cannot generate style-related geometry. 
For example, given a car model and a fire style image in Fig.~\ref{fig:teaser}, two-stage NeRF stylization methods cannot generate a car made of fire with additional geometry representing fire. 
We refer the reader to a recent survey~\citep{chen2023survey} for more techniques on 3D stylization.  

With the rapid development of image-to-3D based on large reconstruction models~\citep{hong2024lrm,wang2024crm,tochilkin2024tripoSR,long2024wonder3d,xu2024dmv3d}, one can also consider generating stylized 3D objects by lifting a single stylized image to 3D. 
However, these pretrained image-to-3D models do not generalize well to stylized images with complex visual effects, e.g., fire around a car, resulting in 3D models with limited geometry quality.
By contrast, our method integrates styles into 3D generation by considering style guidance using reference images in a text-to-3D generation framework. 
Our method follows a single-stage generation principle,  simultaneously optimizing 3D geometry, appearance, and styles.

\section{Background}
\def\s{\mathbf{s}}
\def\z{\mathbf{z}}
\def\x{\mathbf{x}}
\def\I{\mathbf{I}}

\subsection{Text-to-3D generation}
The basic concept in text-to-3D generation is to use a pretrained text-to-image diffusion model to score the rendering of a 3D object described by a text prompt. 
Particularly, given a text prompt $y$ and a pretrained text-to-image diffusion model with the noise prediction network $\epsilon_\phi(\z_t \mid y)$, we aim to generate a 3D object, parameterized by $\theta$, such that its rendering $\x = g(\theta)$ follows the image distribution generated by the pretrained diffusion model. 
This generation can be formulated as an optimization problem with a score distillation sampling (SDS) gradient~\citep{poole2022dreamfusion}: 
\begin{align}
\nabla_{\theta} \mathcal{L}_{SDS} 
&= \mathbb{E}_{t, \epsilon} \left[ \omega(t) \left( \epsilon_{\phi}(\z_t \mid y)- \epsilon \right) \frac{\partial \x}{\partial \theta} \right],
\end{align}
where $\omega(t)$ is a weighting function with $t \sim \mathcal{U}(0.02, 0.098)$, $\epsilon \sim \mathcal{N}(0, I)$, $\z_t = \alpha_t \x + \sigma_t \epsilon$. 
In practice, the score function $\epsilon_{\phi}(\z_t \mid y)$ is implemented with classifier-free guidance~\citep{ho2021classifierfree} to steer the denoising process toward conditional generation to align the generated samples with text prompt $y$.
Several variants of score distillation have been explored to improve the fidelity of generated 3D objects, e.g., variational score distillation (VSD) that expresses the generated objects as probabilistic distributions~\citep{wang2023prolificdreamer}, noise-free score distillation~\citep{katzir2023noisefree} that decomposes distillation scores into interpretable components.

\subsection{Baseline methods for 3D stylization}
\label{sec:baseline}
This section discusses three baseline methods for incorporating style into text-to-3D generation. 
We focus on supporting arbitrary styles in our synthesis, and therefore do not consider techniques that can only support a limited number of styles such as generation guided from style-dependent LoRAs of pretrained diffusion models. 

\paragraph{Style-in-prompt.}
A straightforward baseline method is to use prompt engineering to add a style description directly to the input text prompt. 
For example, instead of having ``ironman'' as the original prompt, we can change to ``golden ironman'' to indicate the desired style of the generation. 
Although this approach can work for simple styles and objects, style representation using text prompts is generally ambiguous and can only capture high-level styles. 
It remains challenging to describe detailed visual elements in styles using text prompts, e.g., styles of sketches.
Empirically, increasing the complexity of text prompts tends to make text-to-3D optimization more challenging to converge. 


\paragraph{Neural style loss.}
The challenges encountered from the first method motivate us to use a reference image to describe detailed visual elements for style transfer. 
Our second baseline method involves using a neural style loss~\citep{gatys2016image} to enforce style consistency between the 3D rendering and a style reference image. 
The style loss is defined by 
\begin{align}
    \mathcal{L}_{style}(\theta) = \| f(\x) - f(\s) \|_2^2,
\end{align}
where $f$ represents the style features extracted by VGG-19~\citep{simonyan2015iclr}. 
Style features were extracted from images using the conv1\_1, conv2\_1, conv3\_1, conv4\_1, conv4\_2, and conv5\_1 layers. 
We apply the style loss as a regularization to an existing score distillation loss.




\paragraph{Textual inversion.}
The third baseline method is specialized for text-to-3D generation by using textual inversion~\citep{gal2022textual,han2023hiper} to map a style reference image to the text embedding of a text-to-image pretrained model, resulting in an augmented text prompt that implicitly encodes the style reference image. 
Particularly, we follow \citep{gal2022textual} to optimize a token $h$ to reproduce the style reference image $\s$ so that the augmented prompt can be defined by $y' = [y + \text{``in the style of''} + h]$. 
We can then use the augmented prompt $y'$ instead of $y$ in a standard text-to-3D optimization.
This baseline method depends on the accuracy of textual inversion that might affect the final 3D generation, and also requires additional computation to perform the textual inversion.

Inspired by the challenges of existing baseline methods, let us now describe our method that aims to circumvent these limitations and generate stylized 3D objects in a robust manner. 

\section{Method}

\subsection{Overview}
Our method seeks a 3D object such that its rendering aligns to an input text prompt and a style reference image.
We optimize a 3D neural representation using score distillation, where the rendering of the 3D object is scored by a pretrained text-to-image diffusion model~\citep{poole2022dreamfusion}. 
Compared to the generic text-to-3D generation, one particular challenge here is to integrate the style reference image into the optimization process to generate stylized 3D objects.
Our method is designed to be a single-stage optimization, where both stylized geometry and appearance are generated simultaneously. 
This differs from some existing 3D stylization methods where only geometry or appearance is optimized to stylize a pre-constructed neural representation~\citep{zhang2022arf,pang2023locally}.

We propose to consume our style reference image using an attention swapping mechanism on the denoising U-Net of the pretrained diffusion model~\citep{jeong2024visual, hertz2023stylealigned} so that the modified diffusion model can generate images analogously to the style reference image. 
We show that this modified pretrained model remains suitable for score distillation, which we then leverage to guide the 3D optimization. 

\subsection{Style-based score distillation}
Mathematically, given a text prompt $y$ and a style reference image $\s$, we seek a 3D object parameterized by $\theta$, with $\x = g(\theta)$ being the rendered image from a differentiable rendering function $g$. 
We apply diffusion on $\x$, with the forward process $q$ and reverse process $p$ as follows. The forward process $q(\z_t \mid \x = g(\theta)) = \mathcal{N}(\alpha_t \x, \sigma_t^2 \I)$ generates a noisy version $\z_t$ of $\x$ at time step $t$ by adding Gaussian noise to $\x$ to remove its structure. The reverse process $p$ predicts the noise from the intermediate state $\z_t$ to reconstruct $\x$. 

We aim to synthesize the 3D object via optimizing its parameter $\theta$ by minimizing the following KL loss:
\begin{align}
    \mathcal{L}(\theta) = KL(q(\z_t \mid \x = g(\theta)) \parallel p_\phi(\z_t \mid y, \s)),
\end{align}
where $p_\phi(\z_t \mid y, \s)$ is a probability distribution with score function parameterized by $\phi$ that conditions on both the text prompt $y$ and the style reference $\s$. 

To model $p_\phi(\z_t \mid y, \s)$, we take inspiration from training-free methods for style transfer using diffusion models~\citep{hertz2023stylealigned,jeong2024visual}.
We assume that there are two denoising processes: one process for generating an image using the original text prompt, and another process for generating a style reference image. 
Here, the style reference image can be generated by its own text prompt, or from textual inversion of a real style image. 
Our goal is to influence the former process so that its generated image has the original content but shares the style in the latter process. 
This can be achieved by sharing features in self-attention blocks~\citep{hertz2023stylealigned} or swapping key and value features at self-attention blocks of the latter process with those of the original process~\citep{jeong2024visual}, allowing features from the style reference images to propagate into image synthesis of the former process. 
As no finetuning is done on the diffusion model itself, this leaves the parameters of the original diffusion model intact, only the score predictions are updated due to the feature changes in the self-attention blocks. 
We adopt this concept for text-to-3D generation as it allows us to use the same pretrained model for original and stylized score distillation for text-to-3D generation.
In our implementation, we follow the swapped attention in visual style prompting~\citep{jeong2024visual} but similar methods such as shared attention~\citep{hertz2023stylealigned} should work as well.

Mathematically, we represent the modified denoising process by a modified score function $\hat{\epsilon}_\phi(\z_t \mid y, \s)$ that shares the same network parameters $\phi$ as the original score function $\epsilon_\phi(\z_t \mid y)$. 
Note that the modified score function has an additional parameter $\s$ which is the style reference image. 
Specifically, assume that the style image can be generated by a prompt $y_s$ so that $\s \sim p_0(\z_0 \mid y_s)$. 
Here we abuse the notation to rewrite the score functions to include self-attention features, namely $\epsilon_\phi(\z_t \mid y; att(y))$ for the original denoising process and $\epsilon_\phi(\z_t \mid y_s; att(y_s))$ for the process generating the style reference image. 
We define the modified score function as
\begin{align}
\hat{\epsilon}_\phi(\z_t \mid y, \s) = \epsilon_\phi(\z_t \mid y; att(y_s)),
\end{align}
where the condition now includes the original prompt $y$ and style features $att(y_s)$.

It is tempting at first to use the modified score function as a standalone distillation for 3D generation, but we very soon realize that this does not work well because the modified score function steers the generated samples toward stylized rendering. 
Predicting 3D shapes from stylized images is highly ambiguous, which often results in low-quality 3D geometry. 
By contrast, the original score function remains useful to steer the denoising process to construct meaningful object shapes. 
This inspires us to propose a combined score function that balances between the original and modified scores, as follows. 

\paragraph{Combined score function.}
We define $p_\phi(\z_t \mid y, \s)$ as a mixture of two distributions in the log space: 
\begin{align}
\log p_\phi(\z_t \mid y, \s) = (1-\lambda) \log p_\phi(\z_t \mid y) + \lambda\log \hat{p}_\phi(\z_t \mid y, \s),
\end{align}
where $p_\phi(\z_t \mid y)$ is the conditional probability distribution of the original pretrained model that only conditions on the text prompt $y$, and $\hat{p}_\phi(\z_t \mid y, \s)$ is the conditional probability distribution of the modified pretrained model that conditions on both the text prompt and the style reference. $\lambda \in [0, 1]$ is the style ratio to control the mixture.
Using this definition, minimizing $KL(q \parallel p_\phi)$ is equivalent to: 
\begin{align}
 \min_{\theta} \mathbb{E}_{\epsilon} & \left[ \log(q(\z_t \mid \x = g(\theta))) \right. \nonumber \\ & \left. - (1 - \lambda) \log(p_\phi(\z_t \mid y)) - \lambda \log(\hat{p}_\phi(\z_t \mid y, \s)) \right].
\end{align}
Taking the derivative w.r.t. $\theta$ results in our stylized score distillation (SSD) gradient: 

\begin{align}
\nabla_{\theta} \mathcal{L}_{SSD} 
&= \mathbb{E}_{t, \epsilon} \left[ \omega(t) \left( (1 - \lambda) \epsilon_{\phi}(\z_t \mid y)  \right.  \right. \nonumber \\ & \left. \left. + \lambda \hat{\epsilon}_{\phi}(\z_t \mid y, \s) - \epsilon \right) \frac{\partial \x}{\partial \theta} \right].
\end{align}

Notably, our stylized score distillation results in linearly interpolated scores of the original and modified pretrained diffusion model, resembling SDS-family gradients. 
This makes extensions on SDS become applicable on our method as well, e.g., classifier-free guidance~\citep{ho2021classifierfree}, and noise-free score distillation~\citep{katzir2023noisefree}, as demonstrated subsequently.

\paragraph{Adaptation to noise-free score distillation.}
Following the noise-free score distillation loss decomposition and taking into account classifier-free guidance~\citep{katzir2023noisefree}, we can represent the score function as a composition of a domain direction $\delta_D$, a noise direction $\delta_N$, and a conditioning direction $\delta_C$. 
The noise-free version of our stylized score distillation can be written as




\begin{align}
\nabla_{\theta} \mathcal{L}_{SNF} &= \mathbb{E}_{t} \left[ \omega(t) \left(
(1 - \lambda) \left(\delta_D + \beta \delta_C \right)  \right.  \right. \nonumber \\ & \left. \left.  + \lambda \left(\hat{\delta}_{D} + \beta \hat{\delta}_{C} \right)
\right) \frac{\partial \x}{\partial \theta} \right],
\end{align}
where the domain directions are defined by
\begin{align}
\delta_D =\begin{cases}
    \epsilon_{\phi}(\z_t \mid y=\emptyset), &\text{if}~~t < 200 \\
    \epsilon_{\phi}(\z_t \mid y=\emptyset) - \epsilon_{\phi}(\z_t \mid y=p_{neg}), & \text{otherwise},\\
\end{cases}
\end{align}
and
\begin{align}
\hat{\delta}_{D} =\begin{cases}
    \hat{\epsilon}_{\phi}(\z_t \mid y=\emptyset, \s), &\text{if}~~t < 200 \\
    \hat{\epsilon}_{\phi}(\z_t \mid y=\emptyset, \s) - \hat{\epsilon}_{\phi}(\z_t \mid y=p_{neg}, \s), & \text{otherwise},\\
\end{cases}
\end{align}
where $p_{neg}$ is a negative prompt to represent out-of-distribution samples such as ``unrealistic, blurry, low quality''. 
The conditioning directions are defined by
\begin{align}
\delta_C = \epsilon_{\phi}(\z_t \mid y) - \epsilon_{\phi}(\z_t \mid y=\emptyset),
\end{align}
and 
\begin{align}
\hat{\delta}_{C} = \hat{\epsilon}_{\phi}(\z_t \mid y, \s) - \hat{\epsilon}_{\phi}(\z_t \mid y=\emptyset, \s),
\end{align}
where $\beta$ is the classifier-free guidance (CFG) scale. 



\subsection{Optimization}
We found that the style ratio $\lambda$ greatly affects the convergence of the optimization as it controls the gradients that steer the denoising process toward generating a generic 3D object and its stylized version. 
We devise a dynamic schedule to adapt the style ratio during optimization as follows. 
We aim for a small style ratio in early iterations so that basic structures in the 3D object can be generated following the vanilla scores. 
In subsequent iterations, we increase the style ratio to favor stylized score distillation, emphasizing the importance of generating stylized 3D objects. 
We explore two dynamic schedules using a square root function:
\begin{align} 
\tau_{\text{sqrt}}(\lambda; \lambda_{\text{max}}, k, K) = \lambda_{\text{max}} \sqrt{\frac{k}{K}},
\end{align} 
and a quadratic function:
\begin{align}
\tau_{\text{quad}}(\lambda; \lambda_{\text{max}}, k, K) = \lambda_{\text{max}} \left(\frac{k}{K}\right)^2 ,
\end{align}
where $k$ and $K$ are the current and total iterations in the optimization, respectively. $\lambda_{\text{max}}$ is the maximum value that the style ratio parameter $\lambda$ will reach at the end of the scaling process when $k = K$.
\section{Experimental Results}

We perform several experiments to demonstrate the effectiveness of our proposed method. First, we compare our method with three baseline methods for stylized text-to-3D generation. 
Second, we demonstrate that our method can be adapted to other score distillation losses.  
Finally, we provide ablation studies to validate the importance of our combined score distillation, as well as perform parameter studies to validate our style ratio scheduling. 
We perform quantitative evaluation of our method through a human-like user study using large language models~\citep{wu2023gpteval3d}.

\begin{figure}[t]
    \centering
    \includegraphics[width=\linewidth]{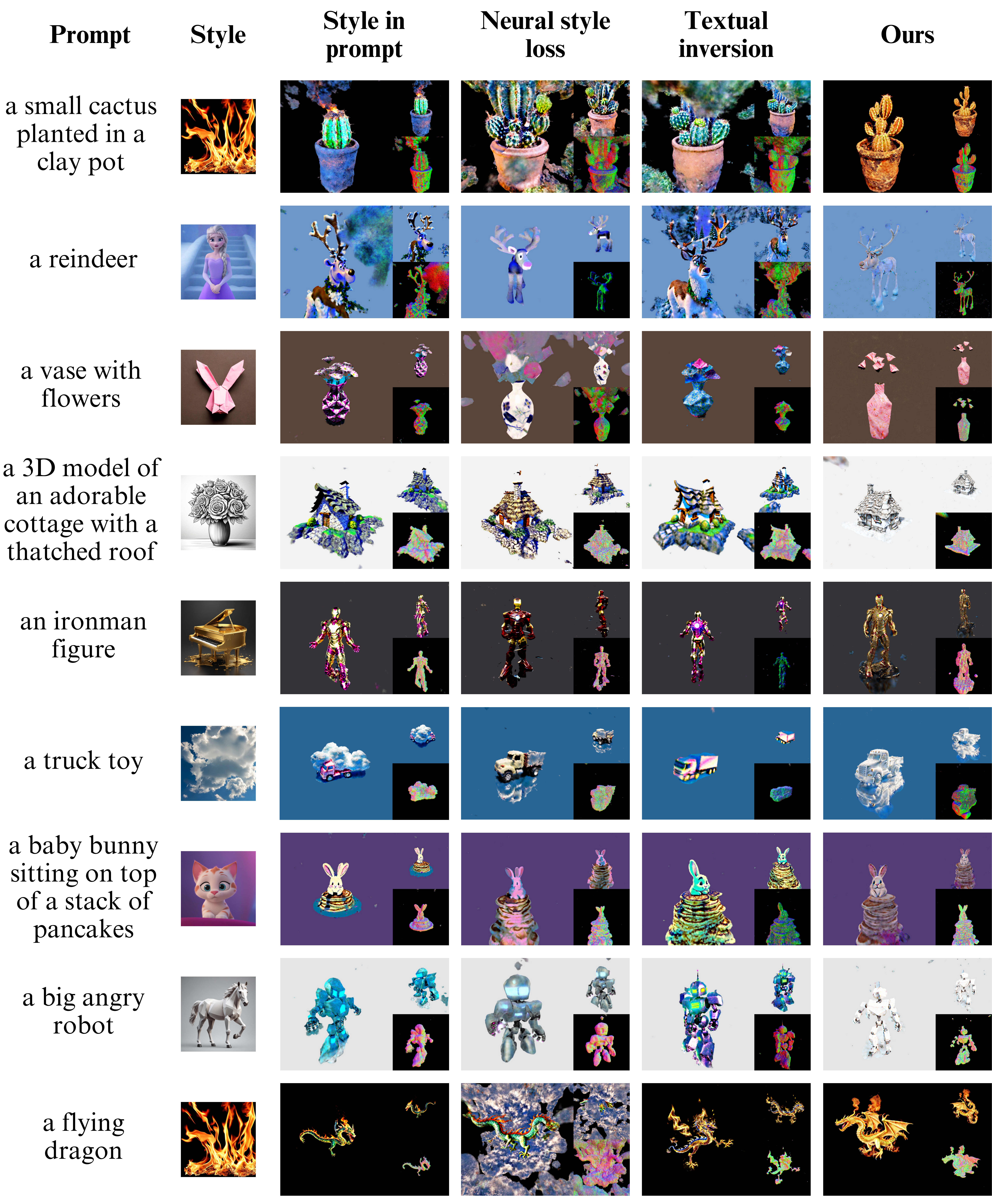}
    \caption{Qualitative comparisons to baseline methods. Our stylized score distillation leads to 3D generation results consistently aligned with the text prompts and the style reference images. Multiple view rendering are shown in the supplementary video.}
    \label{fig:sota_comparison}
\end{figure}

\begin{figure}[t]
    \centering
    \includegraphics[width=0.95\linewidth]{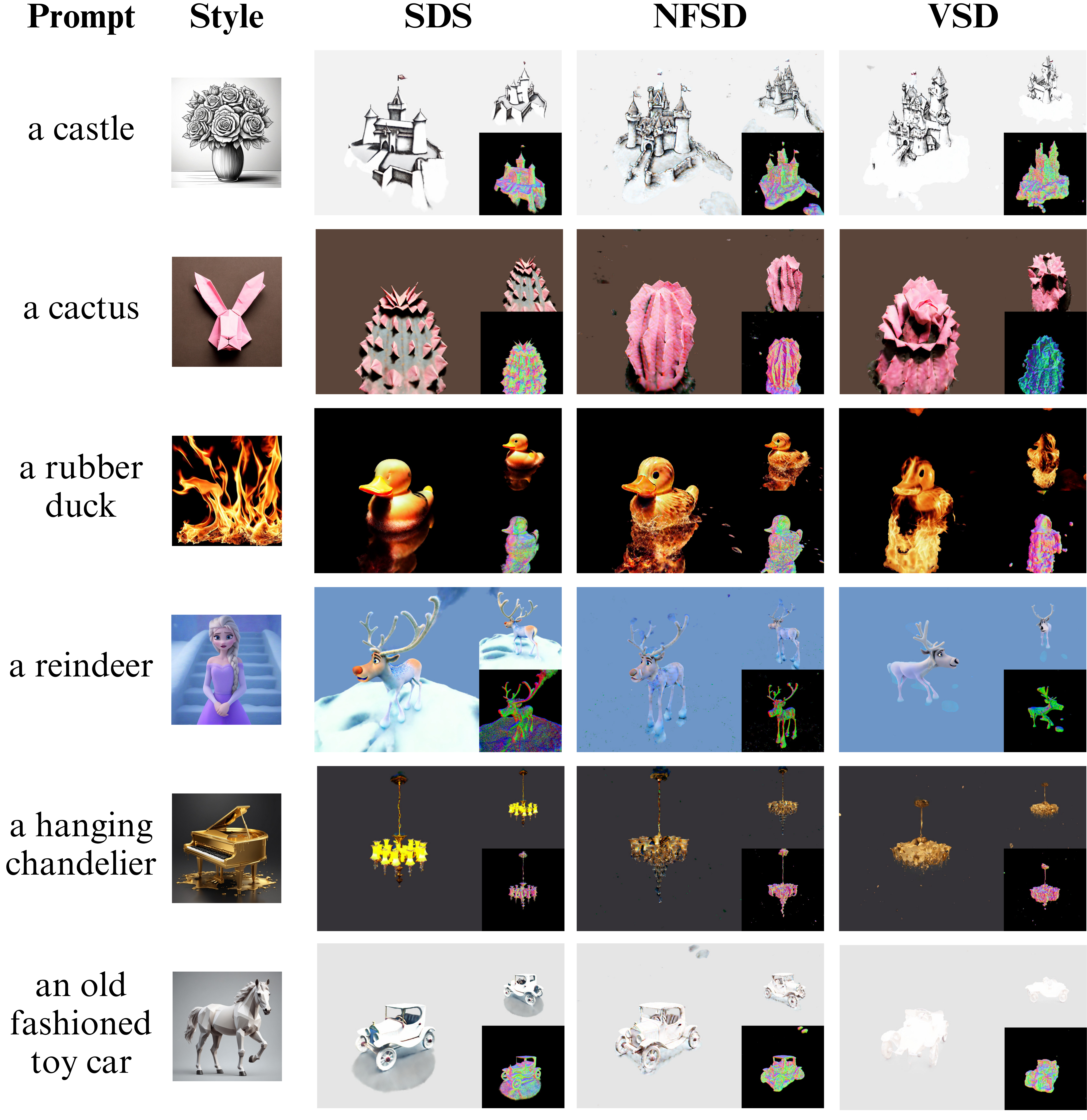}
    \caption{Our method applied to different score distillation losses. Multiple view rendering are shown in the supplementary video.}
    \label{fig:sds_nfsd_vsd}
\end{figure}

\subsection{Implementation details}
We use the implementation of neural radiance fields from threestudio~\citep{threestudio2023}, which is based on NerfAcc~\citep{li2023nerfacc}, as the 3D representation for our optimization. 
We follow the implementation of DreamFusion~\citep{poole2022dreamfusion}, noise-free score distillation~\citep{katzir2023noisefree} and ProlificDreamer~\citep{wang2023prolificdreamer} to implement the score distillations. 
In our method, we apply an augmentation of the text prompt $y$ by concatenating it with a BLIP2 generated caption~\citep{li2023blip2} of the style reference image, and use this augmented prompt for the modified score $\hat{\epsilon}$. 
We set the classifier-free guidance (CFG) scale to 100 for score distillation sampling (SDS)~\citep{poole2022dreamfusion}, and 7.5 for noise-free score distillation (NFSD)~\citep{katzir2023noisefree}, and variational score distillation (VSD)~\citep{wang2023prolificdreamer}.
We use NFSD as the default score distillation for our method.

Our experiments are performed on a NVIDIA RTX 4090 GPU with 24 GB of VRAM. 
Our method optimized a 3D object in approximately 1.5 hours using SDS~\citep{poole2022dreamfusion} or NFSD~\citep{katzir2023noisefree} and 2.5 hours using VSD~\citep{wang2023prolificdreamer}, similar to the training time of the vanilla implementations of these methods.

\subsection{Qualitative results}

We compare our method with the baseline methods proposed in \Cref{sec:baseline}. \Cref{fig:sota_comparison} presents a list of text prompts and style reference images with the corresponding outputs of all methods. 
As can be seen, our results have the best visual quality with consistent alignment to the input pairs of text prompts and reference images. 

\begin{figure}[t!]
    \centering    \includegraphics[width=\linewidth]{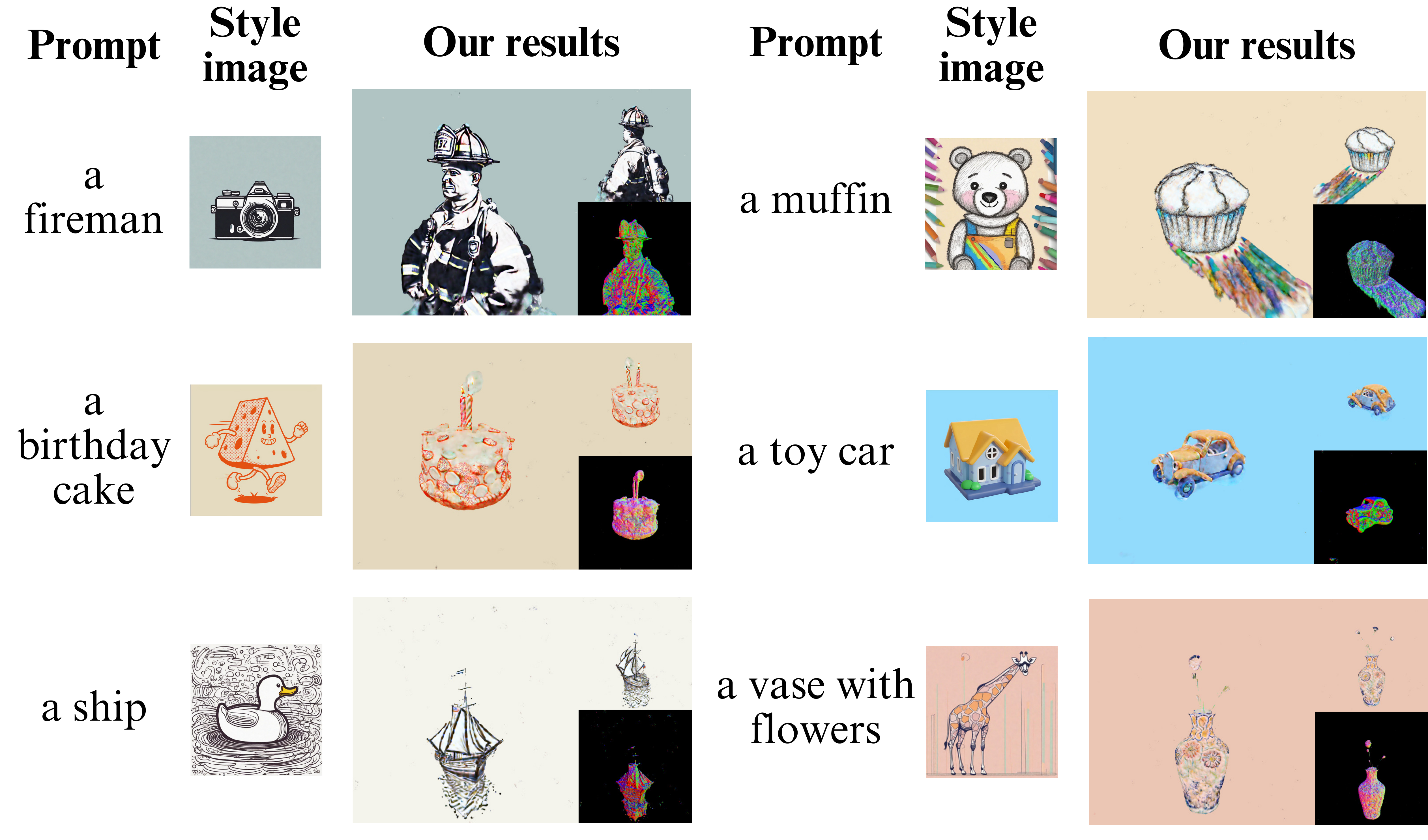}
    \caption{Our results on various style images from ~\citep{hertz2023stylealigned}.}
    \label{fig:more_styles}
\end{figure}

\begin{figure}[t!]
    \centering    \includegraphics[width=\linewidth]{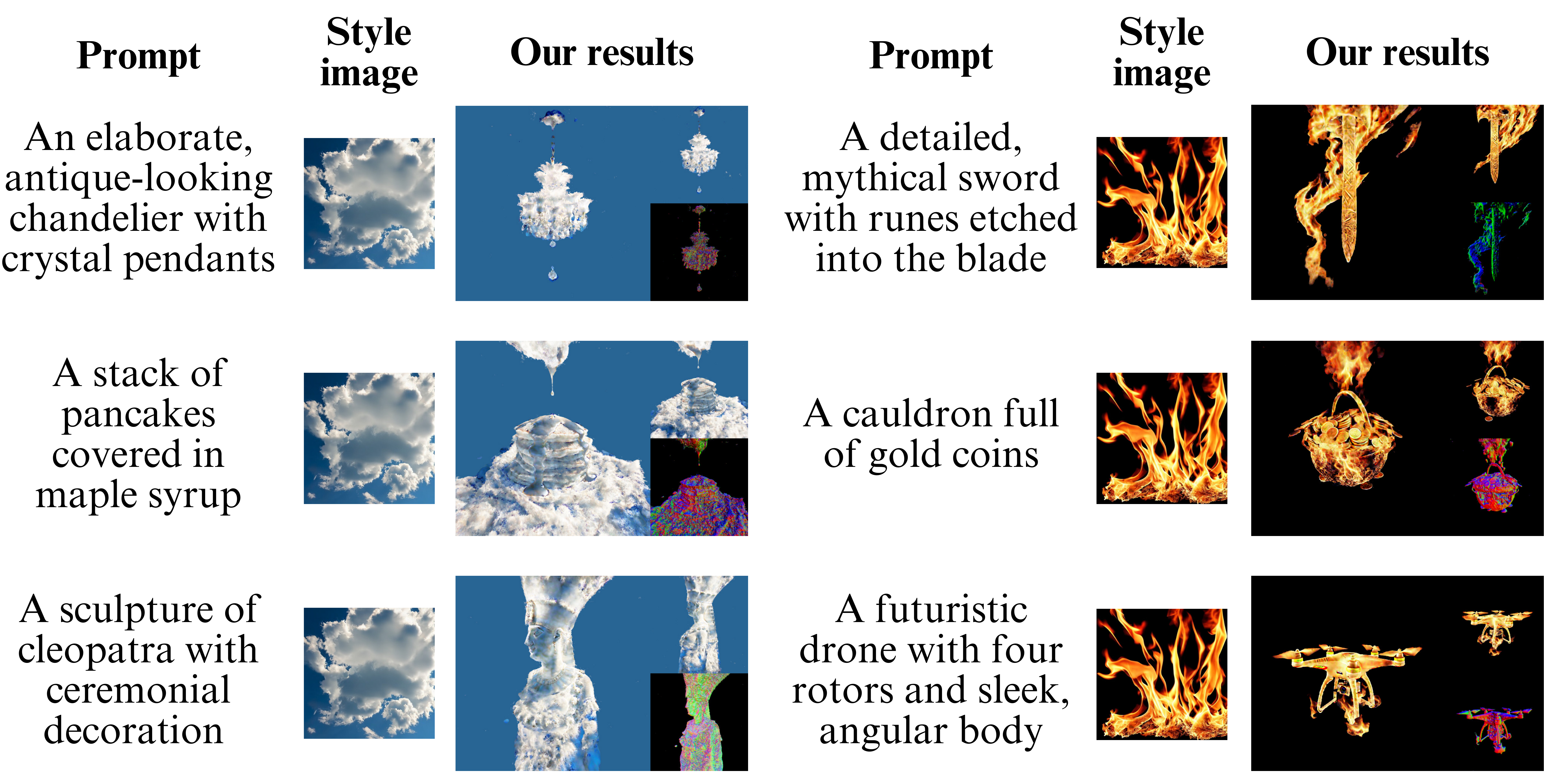}
    \caption{Our results on complex and detailed text prompts.}
    \label{fig:more_complex_prompts}
\end{figure}

To demonstrate the adaptability of our method to other score distillation losses, we apply our method to the vanilla score distillation sampling (SDS), noise-free score distillation (NFSD), and variational score distillation (VSD). 
\Cref{fig:sds_nfsd_vsd} provides the results of our method adapted to these score distillation losses. 
It can be seen that the NFSD variant works best, outperforming SDS and VSD. 
Our method, however, does not yet adapt the LoRA model for scoring noisy rendering in VSD for stylization which is the subject of future work.

To demonstrate the robustness of our method across different styles and complex prompts, we tested it on various styles as in \citep{hertz2023stylealigned}, as shown in \Cref{fig:more_styles}. Additionally, we applied our approach to more complex textual prompts, which are illustrated in \Cref{fig:more_complex_prompts}.

To verify the robustness of our stylized score distillation across training-free methods for style transfer, we attempted our method on StyleAligned \cite{hertz2023stylealigned}. Compared to visual style prompting (VSP)~\citep{jeong2024visual} that only attends to the style image, StyleAligned attends to both the style image and the resulting image. We observed that our SSD with VSP has more natural results. For example, given a fire style image and prompt 'a toy car', SSD with StyleAligned generates fire around a car, while VSP produces a car made of fire (similar trend with their 2D results). A more detailed discussion is provided in the supplementary material.

\subsection{Ablation studies}

\begin{figure}[t!]
    \centering
    \includegraphics[width=\linewidth]{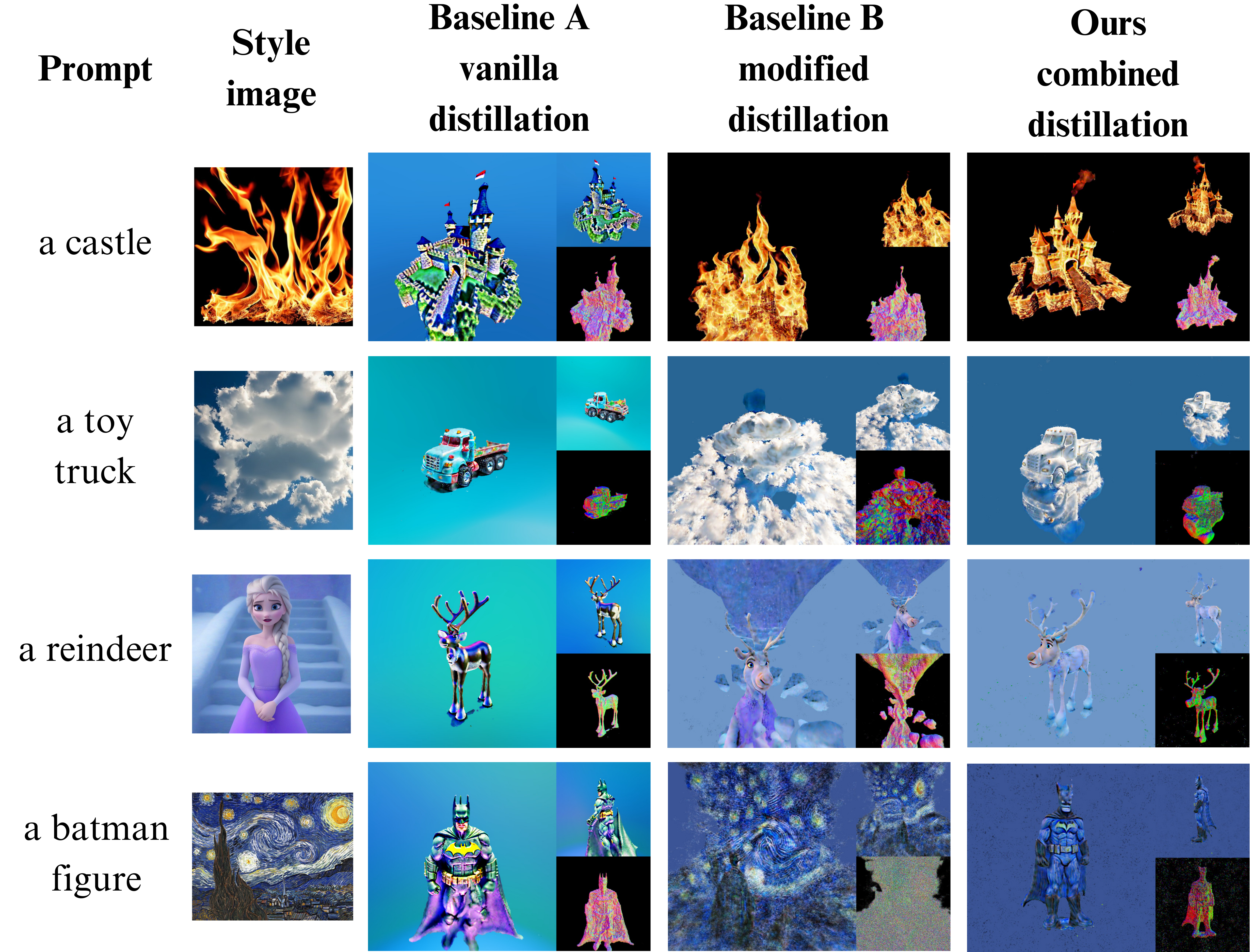}
    \caption{Ablation studies. We confirm the effectiveness of our combined distillation by comparing with two baselines: A) the vanilla text-to-3D generation without styles, B) text-to-3D generation guided by only the modified pretrained model.}
    \label{fig:ablation}
\end{figure}

\begin{figure}[t]
    \centering    \includegraphics[width=\linewidth]{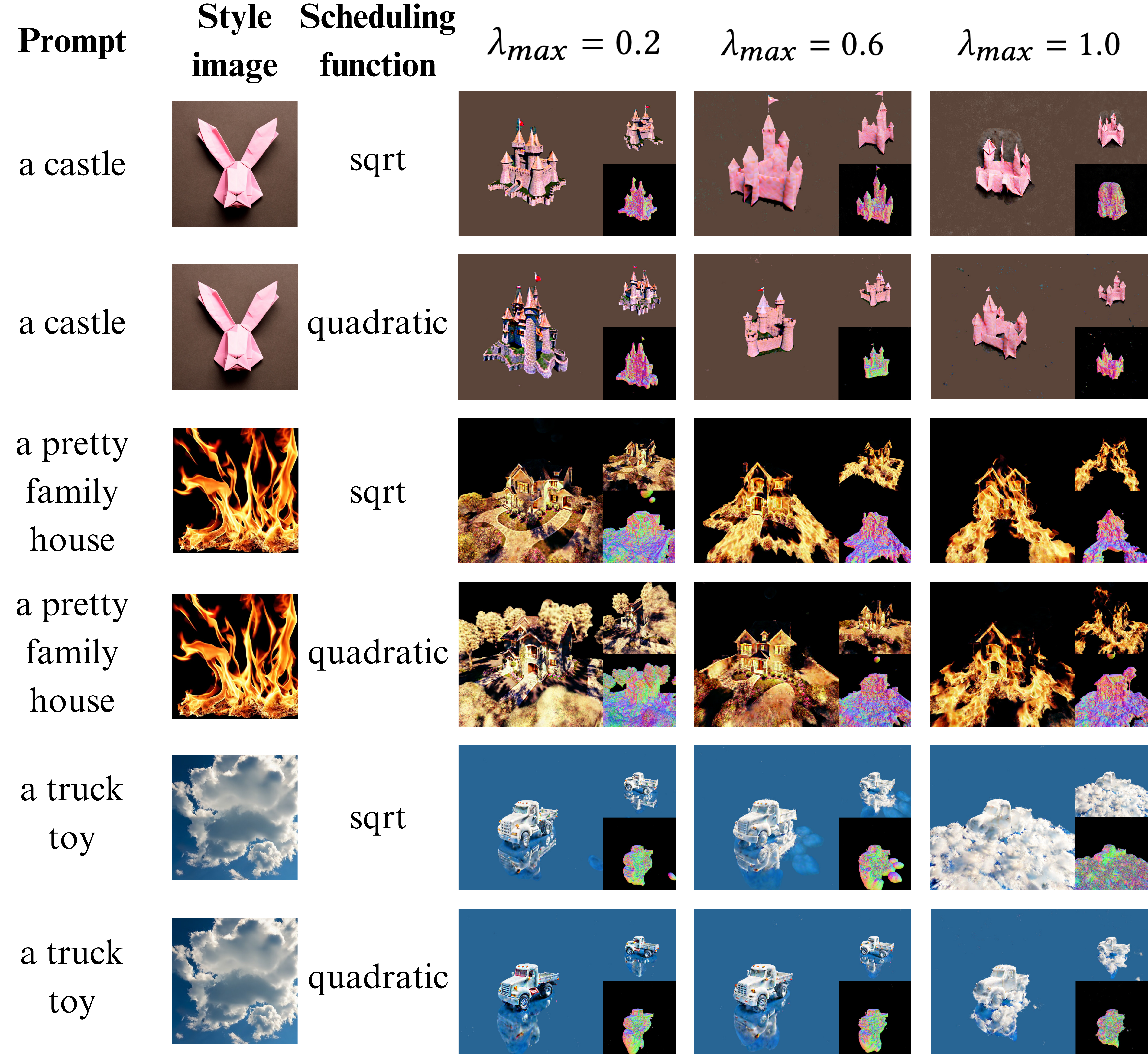}
    \caption{Effects of our schedule function with different style ratios on the stylization results. }
    \label{fig:lambda}
\end{figure}

\Cref{fig:ablation} provides an ablation study on the effectiveness of our method by comparing with two baselines: A) no style reference image (i.e., the vanilla text-to-3D generation), B) text-to-3D generation guided by only the modified pretrained model. 
It is shown that baseline A has high-quality object geometry and appearance, while baseline B has consistent styles with the reference images but causes corrupted geometry. 
This confirms the need for using our stylized score distillation to generate the desired objects in styles.   

\Cref{fig:lambda} provides a study on the choice of the style ratio $\lambda$.  
We found that starting the generation process with an unstyled object and then gradually adding the style works best. 
We evaluated two scheduling functions, quadratic and square root, which sets $\lambda$ between $0$ and $\lambda_{max}$. 
The results are shown in \Cref{fig:lambda}. 
We found that the $sqrt$ schedule works best with $\lambda_{max} = 0.6$ in general. When the style reference image represents abstract concepts without a specific foreground object, it is preferred to use the $quad$ schedule with $\lambda_{max}=1.0$.

\begin{figure}[t]
    \centering    \includegraphics[width=\linewidth]{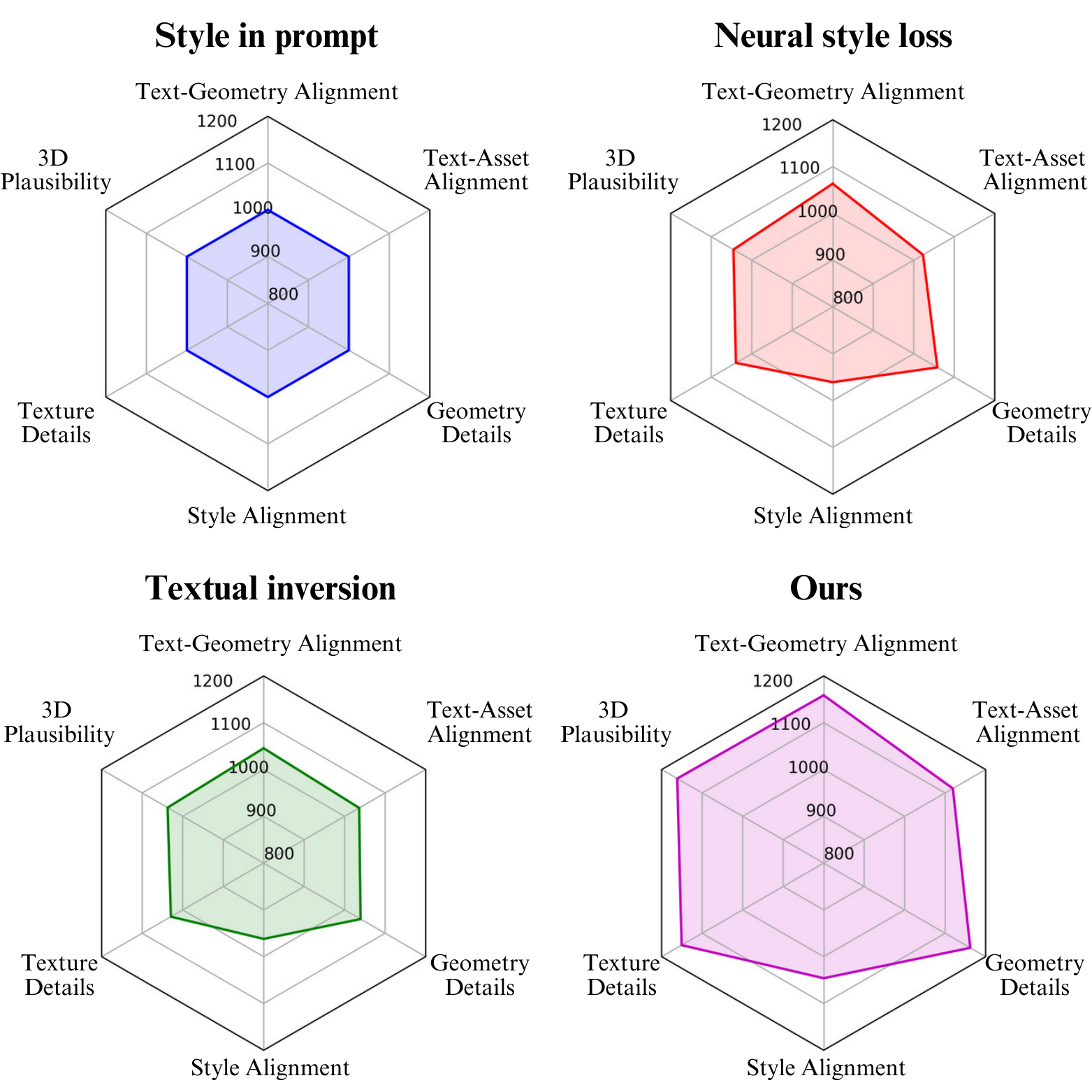}
    \caption{Results of GPTEval3D on text-geometry alignment, text-asset alignment, style alignment, geometry details, texture details, and 3D plausibility, which confirm the effectiveness of our method.}
    \label{fig:gpt_results}
\end{figure}


\subsection{Quantitative results}

The existence of large language models with vision capability (e.g., GPT-4v model~\citep{gpt4v_2023}) allows us to prompt a language model for 3D asset evaluation, which had been demonstrated to work well for the text-to-3D generation task with human-like performance~\citep{wu2023gpteval3d}. 
We follow \citep{wu2023gpteval3d} and extend their GPTEval3D tool to incorporate style evaluation, resulting to six evaluation criteria including text-geometry alignment, text-asset alignment, style alignment, geometry details, texture details, and 3D plausibility.
We use this tool to compare the results generated by four methods, including style-in-prompt, neural style loss, textual inversion, and our method. 
We set up the style-in-prompt method as the base/anchor for the evaluation process. 
We asked GPTEval3D to perform 120 pairwise comparisons, and then calculated the Elo score \citep{elo1966uscf} for each method. The results are presented in \Cref{fig:gpt_results}.
It can be seen that our method outperforms all baseline methods in this evaluation.

\section{Conclusions and limitations}
We present a method for text-to-3D generation in styles. 
Our method is based on a combined score distillation to balance the influence of the original and the modified pretrained diffusion model in generating stylized 3D objects. 
We demonstrated the performance and robustness of our method in various comparisons. 

Our method is not without limitations. 
A particular problem is that our results are prone to the Janus problem~\citep{poole2022dreamfusion}, which could be mitigated by using pretrained diffusion models for multi-view generation~\citep{shi2024MVDream,wang2023imagedream}. 
Additionally, our method is best compatible with SDS-family losses. Adapting our method to non-SDS losses is future work. 
It is of great interest to extend the investigation of stylized text-to-3D generation to videos and 4D data for style references and outputs, respectively. 


\section{Acknowledgments}
This work was conducted with the financial support of the Research Ireland Centre for Research Training in Digitally-Enhanced Reality (d-real) under Grant No. 18/CRT/6224. For the purpose of Open Access, the author has applied a CC BY public copyright licence to any Author Accepted Manuscript version arising from this submission.

This project is supported by Research Ireland under the Research Ireland Frontiers for the Future Programme, award number 22/FFP-P/11522.



\clearpage

{
    \small
    \bibliographystyle{ieeenat_fullname}
    \bibliography{main}

\begin{thebibliography}{82}
\providecommand{\natexlab}[1]{#1}
\providecommand{\url}[1]{\texttt{#1}}
\expandafter\ifx\csname urlstyle\endcsname\relax
  \providecommand{\doi}[1]{doi: #1}\else
  \providecommand{\doi}{doi: \begingroup \urlstyle{rm}\Url}\fi

\bibitem[{Autodesk}(2019)]{maya}
{Autodesk}.
\newblock \emph{Maya}, 2019.

\bibitem[Bahmani et~al.(2024)Bahmani, Skorokhodov, Rong, Wetzstein, Guibas, Wonka, Tulyakov, Park, Tagliasacchi, and Lindell]{bahmani20244dfy}
Sherwin Bahmani, Ivan Skorokhodov, Victor Rong, Gordon Wetzstein, Leonidas Guibas, Peter Wonka, Sergey Tulyakov, Jeong~Joon Park, Andrea Tagliasacchi, and David~B. Lindell.
\newblock 4d-fy: Text-to-4d generation using hybrid score distillation sampling.
\newblock \emph{IEEE Conference on Computer Vision and Pattern Recognition ({CVPR})}, 2024.

\bibitem[{Blender Foundation}(2018)]{blender}
{Blender Foundation}.
\newblock \emph{Blender - a 3D modelling and rendering package}.
\newblock Blender Foundation, Stichting Blender Foundation, Amsterdam, 2018.

\bibitem[Brooks et~al.(2023)Brooks, Holynski, and Efros]{brooks2023instructpix2pix}
Tim Brooks, Aleksander Holynski, and Alexei~A Efros.
\newblock Instructpix2pix: Learning to follow image editing instructions.
\newblock In \emph{Proceedings of the IEEE/CVF Conference on Computer Vision and Pattern Recognition}, pages 18392--18402, 2023.

\bibitem[Chan et~al.(2022)Chan, Lin, Chan, Nagano, Pan, De~Mello, Gallo, Guibas, Tremblay, Khamis, et~al.]{chan2022efficient}
{Eric R} Chan, Connor~Z Lin, Matthew~A Chan, Koki Nagano, Boxiao Pan, Shalini De~Mello, Orazio Gallo, Leonidas~J Guibas, Jonathan Tremblay, Sameh Khamis, et~al.
\newblock Efficient geometry-aware 3d generative adversarial networks.
\newblock In \emph{Proceedings of the IEEE/CVF conference on computer vision and pattern recognition}, pages 16123--16133, 2022.

\bibitem[Chan et~al.(2023)Chan, Nagano, Chan, Bergman, Park, Levy, Aittala, Mello, Karras, and Wetzstein]{chan2023genvs}
Eric~R. Chan, Koki Nagano, Matthew~A. Chan, Alexander~W. Bergman, Jeong~Joon Park, Axel Levy, Miika Aittala, Shalini~De Mello, Tero Karras, and Gordon Wetzstein.
\newblock {GeNVS}: Generative novel view synthesis with {3D}-aware diffusion models.
\newblock In \emph{ICCV}, 2023.

\bibitem[Chen et~al.(2023{\natexlab{a}})Chen, Chen, Jiao, and Jia]{chen2023fantasia3D}
Rui Chen, Yongwei Chen, Ningxin Jiao, and Kui Jia.
\newblock Fantasia3d: Disentangling geometry and appearance for high-quality text-to-3d content creation.
\newblock In \emph{Proceedings of the IEEE/CVF International Conference on Computer Vision (ICCV)}, 2023{\natexlab{a}}.

\bibitem[Chen et~al.(2023{\natexlab{b}})Chen, Shao, Shum, Hua, and Yeung]{chen2023survey}
Yingshu Chen, Guocheng Shao, Ka~Chun Shum, Binh-Son Hua, and Sai-Kit Yeung.
\newblock Advances in 3d neural stylization: A survey.
\newblock \emph{arXiv preprint arXiv:2311.18328}, 2023{\natexlab{b}}.

\bibitem[Chiang et~al.(2022)Chiang, Tsai, Tseng, sheng Lai, and Chiu]{chiang2022stylizing}
Pei-Ze Chiang, Meng-Shiun Tsai, Hung-Yu Tseng, Wei sheng Lai, and Wei-Chen Chiu.
\newblock Stylizing 3d scene via implicit representation and hypernetwork.
\newblock In \emph{WACV}, 2022.

\bibitem[Deitke et~al.(2023{\natexlab{a}})Deitke, Liu, Wallingford, Ngo, Michel, Kusupati, Fan, Laforte, Voleti, Gadre, VanderBilt, Kembhavi, Vondrick, Gkioxari, Ehsani, Schmidt, and Farhadi]{deitke2023objaverseXL}
Matt Deitke, Ruoshi Liu, Matthew Wallingford, Huong Ngo, Oscar Michel, Aditya Kusupati, Alan Fan, Christian Laforte, Vikram Voleti, Samir~Yitzhak Gadre, Eli VanderBilt, Aniruddha Kembhavi, Carl Vondrick, Georgia Gkioxari, Kiana Ehsani, Ludwig Schmidt, and Ali Farhadi.
\newblock Objaverse-{XL}: A universe of 10m+ 3d objects.
\newblock In \emph{Thirty-seventh Conference on Neural Information Processing Systems Datasets and Benchmarks Track}, 2023{\natexlab{a}}.

\bibitem[Deitke et~al.(2023{\natexlab{b}})Deitke, Schwenk, Salvador, Weihs, Michel, VanderBilt, Schmidt, Ehsani, Kembhavi, and Farhadi]{deitke2023objaverse}
Matt Deitke, Dustin Schwenk, Jordi Salvador, Luca Weihs, Oscar Michel, Eli VanderBilt, Ludwig Schmidt, Kiana Ehsani, Aniruddha Kembhavi, and Ali Farhadi.
\newblock Objaverse: A universe of annotated 3d objects.
\newblock In \emph{Proceedings of the IEEE/CVF Conference on Computer Vision and Pattern Recognition}, 2023{\natexlab{b}}.

\bibitem[Elo(1966)]{elo1966uscf}
A.E. Elo.
\newblock \emph{The USCF Rating System: Its Development, Theory, and Applications}.
\newblock United States Chess Federation, 1966.

\bibitem[Fan et~al.(2022)Fan, Jiang, Wang, Gong, Xu, and Wang]{fan2022unified}
Zhiwen Fan, Yifan Jiang, Peihao Wang, Xinyu Gong, Dejia Xu, and Zhangyang Wang.
\newblock Unified implicit neural stylization.
\newblock In \emph{ECCV}, 2022.

\bibitem[Fi\v{s}er et~al.(2016)Fi\v{s}er, Jamri\v{s}ka, Luk\'{a}\v{c}, Shechtman, Asente, Lu, and S\'{y}kora]{fiser2016stylit}
Jakub Fi\v{s}er, Ond\v{r}ej Jamri\v{s}ka, Michal Luk\'{a}\v{c}, Eli Shechtman, Paul Asente, Jingwan Lu, and Daniel S\'{y}kora.
\newblock Stylit: Illumination-guided example-based stylization of 3d renderings.
\newblock \emph{ACM Trans. Graph.}, 2016.

\bibitem[Gal et~al.(2022)Gal, Patashnik, Maron, Bermano, Chechik, and Cohen-Or]{gal2022stylegan}
Rinon Gal, Or Patashnik, Haggai Maron, Amit~H Bermano, Gal Chechik, and Daniel Cohen-Or.
\newblock Stylegan-nada: Clip-guided domain adaptation of image generators.
\newblock \emph{ACM Transactions on Graphics (TOG)}, 2022.

\bibitem[Gal et~al.(2023)Gal, Alaluf, Atzmon, Patashnik, Bermano, Chechik, and Cohen-Or]{gal2022textual}
Rinon Gal, Yuval Alaluf, Yuval Atzmon, Or Patashnik, Amit~H. Bermano, Gal Chechik, and Daniel Cohen-Or.
\newblock An image is worth one word: Personalizing text-to-image generation using textual inversion.
\newblock In \emph{ICLR}, 2023.

\bibitem[Gatys et~al.(2016)Gatys, Ecker, and Bethge]{gatys2016image}
{Leon A} Gatys, {Alexander S} Ecker, and Matthias Bethge.
\newblock Image style transfer using convolutional neural networks.
\newblock In \emph{Proceedings of the IEEE conference on computer vision and pattern recognition}, pages 2414--2423, 2016.

\bibitem[Goodfellow et~al.(2014)Goodfellow, Pouget-Abadie, Mirza, Xu, Warde-Farley, Ozair, Courville, and Bengio]{goodfellow2014generative}
Ian Goodfellow, Jean Pouget-Abadie, Mehdi Mirza, Bing Xu, David Warde-Farley, Sherjil Ozair, Aaron Courville, and Yoshua Bengio.
\newblock Generative adversarial nets.
\newblock In \emph{Advances in neural information processing systems}, 2014.

\bibitem[Guo et~al.(2023)Guo, Liu, Shao, Laforte, Voleti, Luo, Chen, Zou, Wang, Cao, and Zhang]{threestudio2023}
Yuan-Chen Guo, Ying-Tian Liu, Ruizhi Shao, Christian Laforte, Vikram Voleti, Guan Luo, Chia-Hao Chen, Zi-Xin Zou, Chen Wang, Yan-Pei Cao, and Song-Hai Zhang.
\newblock threestudio: A unified framework for 3d content generation.
\newblock \url{https://github.com/threestudio-project/threestudio}, 2023.

\bibitem[Han et~al.(2023)Han, Yang, Kwon, and Ye]{han2023hiper}
Inhwa Han, Serin Yang, Taesung Kwon, and Jong~Chul Ye.
\newblock Highly personalized text embedding for image manipulation by stable diffusion.
\newblock \emph{arXiv preprint arXiv:2303.08767}, 2023.

\bibitem[Hertz et~al.(2023)Hertz, Aberman, and Cohen-Or]{hertz2023delta}
Amir Hertz, Kfir Aberman, and Daniel Cohen-Or.
\newblock Delta denoising score.
\newblock In \emph{Proceedings of the IEEE/CVF International Conference on Computer Vision}, pages 2328--2337, 2023.

\bibitem[Hertz et~al.(2024)Hertz, Voynov, Fruchter, and Cohen-Or]{hertz2023stylealigned}
Amir Hertz, Andrey Voynov, Shlomi Fruchter, and Daniel Cohen-Or.
\newblock Style aligned image generation via shared attention.
\newblock In \emph{CVPR}, 2024.

\bibitem[Hertzmann et~al.(2001)Hertzmann, Jacobs, Oliver, Curless, and Salesin]{hertzmann2001analogies}
Aaron Hertzmann, Charles~E Jacobs, Nuria Oliver, Brian Curless, and David~H Salesin.
\newblock Image analogies.
\newblock In \emph{Proceedings of the 28th annual conference on Computer graphics and interactive techniques}, pages 327--340, 2001.

\bibitem[Ho and Salimans(2021)]{ho2021classifierfree}
Jonathan Ho and Tim Salimans.
\newblock Classifier-free diffusion guidance.
\newblock \emph{NeurIPS 2021 Workshop on Deep Generative Models and Downstream Applications}, 2021.

\bibitem[Ho et~al.(2020)Ho, Jain, and Abbeel]{ho2020denoising}
Jonathan Ho, Ajay Jain, and Pieter Abbeel.
\newblock Denoising diffusion probabilistic models.
\newblock In \emph{Advances in neural information processing systems}, pages 6840--6851, 2020.

\bibitem[Hong et~al.(2024)Hong, Zhang, Gu, Bi, Zhou, Liu, Liu, Sunkavalli, Bui, and Tan]{hong2024lrm}
Yicong Hong, Kai Zhang, Jiuxiang Gu, Sai Bi, Yang Zhou, Difan Liu, Feng Liu, Kalyan Sunkavalli, Trung Bui, and Hao Tan.
\newblock Lrm: Large reconstruction model for single image to 3d.
\newblock In \emph{ICLR}, 2024.

\bibitem[Huang and Belongie(2017)]{huang2017arbitrary}
Xun Huang and Serge Belongie.
\newblock Arbitrary style transfer in real-time with adaptive instance normalization.
\newblock In \emph{Proceedings of the IEEE international conference on computer vision}, pages 1501--1510, 2017.

\bibitem[Huang et~al.(2022)Huang, He, Yuan, Lai, and Gao]{huang2022stylizedNeRF}
Yi-Hua Huang, Yue He, Yu-Jie Yuan, Yu-Kun Lai, and Lin Gao.
\newblock Stylizednerf: Consistent 3d scene stylization as stylized nerf via 2d-3d mutual learning.
\newblock In \emph{CVPR}, 2022.

\bibitem[Isola et~al.(2017)Isola, Zhu, Zhou, and Efros]{isola2017image}
Phillip Isola, Jun-Yan Zhu, Tinghui Zhou, and Alexei~A Efros.
\newblock Image-to-image translation with conditional adversarial networks.
\newblock In \emph{Proceedings of the IEEE conference on computer vision and pattern recognition}, pages 1125--1134, 2017.

\bibitem[Jeong et~al.(2024)Jeong, Kim, Choi, Lee, and Uh]{jeong2024visual}
Jaeseok Jeong, Junho Kim, Yunjey Choi, Gayoung Lee, and Youngjung Uh.
\newblock Visual style prompting with swapping self-attention.
\newblock \emph{arXiv preprint arXiv:2402.12974}, 2024.

\bibitem[Jing et~al.(2019)Jing, Yang, Feng, Ye, Yu, and Song]{jing2019neural}
Yongcheng Jing, Yezhou Yang, Zunlei Feng, Jingwen Ye, Yizhou Yu, and Mingli Song.
\newblock Neural style transfer: A review.
\newblock \emph{IEEE transactions on visualization and computer graphics}, 2019.

\bibitem[Johnson et~al.(2016)Johnson, Alahi, and Fei-Fei]{johnson2016perceptual}
Justin Johnson, Alexandre Alahi, and Li Fei-Fei.
\newblock Perceptual losses for real-time style transfer and super-resolution.
\newblock In \emph{European conference on computer vision}, pages 694--711. Springer, 2016.

\bibitem[Karras et~al.(2019)Karras, Laine, and Aila]{karras2019style}
Tero Karras, Samuli Laine, and Timo Aila.
\newblock A style-based generator architecture for generative adversarial networks.
\newblock In \emph{Proceedings of the IEEE/CVF conference on computer vision and pattern recognition}, pages 4401--4410, 2019.

\bibitem[Katzir et~al.(2024)Katzir, Patashnik, Cohen-Or, and Lischinski]{katzir2023noisefree}
Oren Katzir, Or Patashnik, Daniel Cohen-Or, and Dani Lischinski.
\newblock Noise-free score distillation.
\newblock \emph{ICLR}, 2024.

\bibitem[Koo et~al.(2024)Koo, Park, and Sung]{koo2024posterior}
Juil Koo, Chanho Park, and Minhyuk Sung.
\newblock Posterior distillation sampling.
\newblock In \emph{CVPR}, 2024.

\bibitem[Kwon and Ye(2022)]{kwon2022clipstyler}
Gihyun Kwon and Jong~Chul Ye.
\newblock Clipstyler: Image style transfer with a single text condition.
\newblock In \emph{Proceedings of the IEEE/CVF Conference on Computer Vision and Pattern Recognition}, pages 18062--18071, 2022.

\bibitem[Li et~al.(2023{\natexlab{a}})Li, Xue, Liu, and Lai]{li2023bbdm}
Bo Li, Kaitao Xue, Bin Liu, and Yu-Kun Lai.
\newblock Bbdm: Image-to-image translation with brownian bridge diffusion models.
\newblock In \emph{Proceedings of the IEEE/CVF Conference on Computer Vision and Pattern Recognition}, pages 1952--1961, 2023{\natexlab{a}}.

\bibitem[Li et~al.(2023{\natexlab{b}})Li, Li, Savarese, and Hoi]{li2023blip2}
Junnan Li, Dongxu Li, Silvio Savarese, and Steven Hoi.
\newblock {BLIP}-2: Bootstrapping language-image pre-training with frozen image encoders and large language models.
\newblock In \emph{Proceedings of the 40th International Conference on Machine Learning}, pages 19730--19742, 2023{\natexlab{b}}.

\bibitem[Li et~al.(2023{\natexlab{c}})Li, Gao, Tancik, and Kanazawa]{li2023nerfacc}
Ruilong Li, Hang Gao, Matthew Tancik, and Angjoo Kanazawa.
\newblock Nerfacc: Efficient sampling accelerates nerfs, 2023{\natexlab{c}}.

\bibitem[Li et~al.(2019)Li, Liu, Kautz, and Yang]{li2019linear}
Xueting Li, Sifei Liu, Jan Kautz, and Ming-Hsuan Yang.
\newblock Learning linear transformations for fast image and video style transfer.
\newblock In \emph{Proceedings of the IEEE/CVF Conference on Computer Vision and Pattern Recognition}, pages 3809--3817, 2019.

\bibitem[Liao et~al.(2017)Liao, Yao, Yuan, Hua, and Kang]{liao2017analogies}
Jing Liao, Yuan Yao, Lu Yuan, Gang Hua, and Sing~Bing Kang.
\newblock Visual attribute transfer through deep image analogy.
\newblock \emph{ACM Trans. Graph.}, 2017.

\bibitem[Lin et~al.(2023)Lin, Gao, Tang, Takikawa, Zeng, Huang, Kreis, Fidler, Liu, and Lin]{lin2023magic3d}
{Chen-Hsuan} Lin, Jun Gao, Luming Tang, Towaki Takikawa, Xiaohui Zeng, Xun Huang, Karsten Kreis, Sanja Fidler, Ming-Yu Liu, and Tsung-Yi Lin.
\newblock Magic3d: High-resolution text-to-3d content creation.
\newblock In \emph{Proceedings of the IEEE/CVF Conference on Computer Vision and Pattern Recognition}, 2023.

\bibitem[Liu et~al.(2023{\natexlab{a}})Liu, Zhan, Chen, Zhang, Yu, Saddik, Lu, and Xing]{liu2023stylerf}
Kunhao Liu, Fangneng Zhan, Yiwen Chen, Jiahui Zhang, Yingchen Yu, Abdulmotaleb~El Saddik, Shijian Lu, and Eric Xing.
\newblock Stylerf: Zero-shot 3d style transfer of neural radiance fields.
\newblock In \emph{CVPR}, 2023{\natexlab{a}}.

\bibitem[Liu et~al.(2017)Liu, Breuel, and Kautz]{liu2017unsupervised}
{Ming-Yu} Liu, Thomas Breuel, and Jan Kautz.
\newblock Unsupervised image-to-image translation networks.
\newblock In \emph{Advances in neural information processing systems}, pages 700--708, 2017.

\bibitem[Liu et~al.(2023{\natexlab{b}})Liu, Wu, Hoorick, Tokmakov, Zakharov, and Vondrick]{liu2023zero1to3}
Ruoshi Liu, Rundi Wu, Basile~Van Hoorick, Pavel Tokmakov, Sergey Zakharov, and Carl Vondrick.
\newblock Zero-1-to-3: Zero-shot one image to 3d object.
\newblock In \emph{ICCV}, 2023{\natexlab{b}}.

\bibitem[Long et~al.(2024)Long, Guo, Lin, Liu, Dou, Liu, Ma, Zhang, Habermann, Theobalt, et~al.]{long2024wonder3d}
Xiaoxiao Long, Yuan-Chen Guo, Cheng Lin, Yuan Liu, Zhiyang Dou, Lingjie Liu, Yuexin Ma, Song-Hai Zhang, Marc Habermann, Christian Theobalt, et~al.
\newblock Wonder3d: Single image to 3d using cross-domain diffusion.
\newblock In \emph{CVPR}, 2024.

\bibitem[Lorraine et~al.(2023)Lorraine, Xie, Zeng, Lin, Takikawa, Sharp, Lin, Liu, Fidler, and Lucas]{lorraine2023att3d}
Jonathan Lorraine, Kevin Xie, Xiaohui Zeng, Chen-Hsuan Lin, Towaki Takikawa, Nicholas Sharp, Tsung-Yi Lin, Ming-Yu Liu, Sanja Fidler, and James Lucas.
\newblock Att3d: Amortized text-to-3d object synthesis.
\newblock \emph{ICCV}, 2023.

\bibitem[Meng et~al.(2022)Meng, He, Song, Song, Wu, Zhu, and Ermon]{meng2021sdedit}
Chenlin Meng, Yutong He, Yang Song, Jiaming Song, Jiajun Wu, Jun-Yan Zhu, and Stefano Ermon.
\newblock {SDE}dit: Guided image synthesis and editing with stochastic differential equations.
\newblock In \emph{International Conference on Learning Representations}, 2022.

\bibitem[Metzer et~al.(2023)Metzer, Richardson, Patashnik, Giryes, and Cohen-Or]{metzer2023latent}
Gal Metzer, Elad Richardson, Or Patashnik, Raja Giryes, and Daniel Cohen-Or.
\newblock Latent-nerf for shape-guided generation of 3d shapes and textures.
\newblock In \emph{Proceedings of the IEEE/CVF Conference on Computer Vision and Pattern Recognition}, pages 12663--12673, 2023.

\bibitem[Mildenhall et~al.(2020)Mildenhall, Srinivasan, Tancik, Barron, Ramamoorthi, and Ng]{mildenhall2020nerf}
Ben Mildenhall, Pratul~P Srinivasan, Matthew Tancik, Jonathan~T Barron, Ravi Ramamoorthi, and Ren Ng.
\newblock Nerf: Representing scenes as neural radiance fields for view synthesis.
\newblock In \emph{European conference on computer vision}, pages 405--421. Springer, 2020.

\bibitem[Nguyen-Phuoc et~al.(2022)Nguyen-Phuoc, Liu, and Xiao]{nguyen2022snerf}
Thu Nguyen-Phuoc, Feng Liu, and Lei Xiao.
\newblock Snerf: stylized neural implicit representations for 3d scenes.
\newblock \emph{ACM Transactions on Graphics}, 2022.

\bibitem[Nichol et~al.(2022)Nichol, Jun, Dhariwal, Mishkin, and Chen]{nichol2022point}
Alex Nichol, Heewoo Jun, Prafulla Dhariwal, Pamela Mishkin, and Mark Chen.
\newblock Point-e: A system for generating 3d point clouds from complex prompts.
\newblock \emph{arXiv 2212.08751}, 2022.

\bibitem[OpenAI(2023{\natexlab{a}})]{dalle3_2023}
OpenAI.
\newblock Dall·e 3, 2023{\natexlab{a}}.

\bibitem[OpenAI(2023{\natexlab{b}})]{gpt4v_2023}
OpenAI.
\newblock Gpt-4v, 2023{\natexlab{b}}.

\bibitem[Pang et~al.(2023)Pang, Hua, and Yeung]{pang2023locally}
{Hong-Wing} Pang, Binh-Son Hua, and Sai-Kit Yeung.
\newblock Locally stylized neural radiance fields.
\newblock In \emph{ICCV}, 2023.

\bibitem[Parmar et~al.(2023)Parmar, Kumar~Singh, Zhang, Li, Lu, and Zhu]{parmar2023zero}
Gaurav Parmar, Krishna Kumar~Singh, Richard Zhang, Yijun Li, Jingwan Lu, and Jun-Yan Zhu.
\newblock Zero-shot image-to-image translation.
\newblock In \emph{ACM SIGGRAPH 2023 Conference Proceedings}, pages 1--11, 2023.

\bibitem[Patashnik et~al.(2021)Patashnik, Wu, Shechtman, Cohen-Or, and Lischinski]{patashnik2021styleclip}
Or Patashnik, Zongze Wu, Eli Shechtman, Daniel Cohen-Or, and Dani Lischinski.
\newblock Styleclip: Text-driven manipulation of stylegan imagery.
\newblock In \emph{Proceedings of the IEEE/CVF International Conference on Computer Vision}, pages 2085--2094, 2021.

\bibitem[Po et~al.(2024)Po, Yifan, Golyanik, Aberman, Barron, Bermano, Chan, Dekel, Holynski, Kanazawa, Liu, Liu, Mildenhall, Nießner, Ommer, Theobalt, Wonka, and Wetzstein]{po2023survey}
Ryan Po, Wang Yifan, Vladislav Golyanik, Kfir Aberman, Jonathan~T. Barron, Amit~H. Bermano, Eric~Ryan Chan, Tali Dekel, Aleksander Holynski, Angjoo Kanazawa, C.~Karen Liu, Lingjie Liu, Ben Mildenhall, Matthias Nießner, Björn Ommer, Christian Theobalt, Peter Wonka, and Gordon Wetzstein.
\newblock State of the art on diffusion models for visual computing.
\newblock \emph{Computer Graphics Forum}, 2024.

\bibitem[Poole et~al.(2023)Poole, Jain, Barron, and Mildenhall]{poole2022dreamfusion}
Ben Poole, Ajay Jain, Jonathan~T. Barron, and Ben Mildenhall.
\newblock Dreamfusion: Text-to-3d using 2d diffusion.
\newblock In \emph{The Eleventh International Conference on Learning Representations}, 2023.

\bibitem[Roeder et~al.(2017)Roeder, Wu, and Duvenaud]{roeder2017sticking}
Geoffrey Roeder, Yuhuai Wu, and David Duvenaud.
\newblock Sticking the landing: Simple, lower-variance gradient estimators for variational inference, 2017.

\bibitem[Rombach et~al.(2022)Rombach, Blattmann, Lorenz, Esser, and Ommer]{rombach2022high}
Robin Rombach, Andreas Blattmann, Dominik Lorenz, Patrick Esser, and Bj{\"o}rn Ommer.
\newblock High-resolution image synthesis with latent diffusion models.
\newblock In \emph{Proceedings of the IEEE/CVF conference on computer vision and pattern recognition}, pages 10684--10695, 2022.

\bibitem[Ruiz et~al.(2023)Ruiz, Li, Jampani, Pritch, Rubinstein, and Aberman]{ruiz2022dreambooth}
Nataniel Ruiz, Yuanzhen Li, Varun Jampani, Yael Pritch, Michael Rubinstein, and Kfir Aberman.
\newblock Dreambooth: Fine tuning text-to-image diffusion models for subject-driven generation.
\newblock In \emph{CVPR}, 2023.

\bibitem[Shi et~al.(2024)Shi, Wang, Ye, Mai, Li, and Yang]{shi2024MVDream}
Yichun Shi, Peng Wang, Jianglong Ye, Long Mai, Kejie Li, and Xiao Yang.
\newblock Mvdream: Multi-view diffusion for 3d generation.
\newblock In \emph{ICLR}, 2024.

\bibitem[Simonyan and Zisserman(2015)]{simonyan2015iclr}
Karen Simonyan and Andrew Zisserman.
\newblock Very deep convolutional networks for large-scale image recognition.
\newblock In \emph{International Conference on Learning Representations}, 2015.

\bibitem[Su et~al.(2023)Su, Song, Meng, and Ermon]{su2022dual}
Xuan Su, Jiaming Song, Chenlin Meng, and Stefano Ermon.
\newblock Dual diffusion implicit bridges for image-to-image translation.
\newblock In \emph{The Eleventh International Conference on Learning Representations}, 2023.

\bibitem[S\'{y}kora et~al.(2019)S\'{y}kora, Jamri\v{s}ka, Texler, Fi\v{s}er, Luk\'{a}\v{c}, Lu, and Shechtman]{sykora2019styleblit}
Daniel S\'{y}kora, Ond\v{r}ej Jamri\v{s}ka, Ond\v{r}ej Texler, Jakub Fi\v{s}er, Michal Luk\'{a}\v{c}, Jingwan Lu, and Eli Shechtman.
\newblock {StyleBlit}: Fast example-based stylization with local guidance.
\newblock \emph{Computer Graphics Forum}, 38\penalty0 (2):\penalty0 83--91, 2019.

\bibitem[Szymanowicz et~al.(2023)Szymanowicz, Rupprecht, and Vedaldi]{szymanowicz23viewset}
Stanislaw Szymanowicz, Christian Rupprecht, and Andrea Vedaldi.
\newblock Viewset diffusion: (0-)image-conditioned {3D} generative models from {2D} data.
\newblock In \emph{ICCV}, 2023.

\bibitem[Tang et~al.(2024)Tang, Ren, Zhou, Liu, and Zeng]{tang2024dreamgaussian}
Jiaxiang Tang, Jiawei Ren, Hang Zhou, Ziwei Liu, and Gang Zeng.
\newblock Dreamgaussian: Generative gaussian splatting for efficient 3d content creation.
\newblock \emph{ICLR}, 2024.

\bibitem[Tochilkin et~al.(2024)Tochilkin, Pankratz, Liu, Huang, , Letts, Li, Liang, Laforte, Jampani, and Cao]{tochilkin2024tripoSR}
Dmitry Tochilkin, David Pankratz, Zexiang Liu, Zixuan Huang, , Adam Letts, Yangguang Li, Ding Liang, Christian Laforte, Varun Jampani, and Yan-Pei Cao.
\newblock Triposr: Fast 3d object reconstruction from a single image.
\newblock \emph{arXiv preprint arXiv:2403.02151}, 2024.

\bibitem[Wang et~al.(2023{\natexlab{a}})Wang, Du, Li, Yeh, and Shakhnarovich]{wang2023score}
Haochen Wang, Xiaodan Du, Jiahao Li, Raymond~A Yeh, and Greg Shakhnarovich.
\newblock Score jacobian chaining: Lifting pretrained 2d diffusion models for 3d generation.
\newblock In \emph{Proceedings of the IEEE/CVF Conference on Computer Vision and Pattern Recognition}, 2023{\natexlab{a}}.

\bibitem[Wang and Shi(2023)]{wang2023imagedream}
Peng Wang and Yichun Shi.
\newblock Imagedream: Image-prompt multi-view diffusion for 3d generation.
\newblock \emph{arXiv preprint arXiv:2312.02201}, 2023.

\bibitem[Wang et~al.(2023{\natexlab{b}})Wang, Lu, Wang, Bao, Li, Su, and Zhu]{wang2023prolificdreamer}
Zhengyi Wang, Cheng Lu, Yikai Wang, Fan Bao, Chongxuan Li, Hang Su, and Jun Zhu.
\newblock Prolificdreamer: High-fidelity and diverse text-to-3d generation with variational score distillation.
\newblock In \emph{Advances in Neural Information Processing Systems}, 2023{\natexlab{b}}.

\bibitem[Wang et~al.(2024)Wang, Wang, Chen, Xiang, Chen, Yu, Li, Su, and Zhu]{wang2024crm}
Zhengyi Wang, Yikai Wang, Yifei Chen, Chendong Xiang, Shuo Chen, Dajiang Yu, Chongxuan Li, Hang Su, and Jun Zhu.
\newblock Crm: Single image to 3d textured mesh with convolutional reconstruction model.
\newblock In \emph{ECCV}, 2024.

\bibitem[Wu et~al.(2024)Wu, Yang, Li, Zhang, Liu, Guibas, Lin, and Wetzstein]{wu2023gpteval3d}
Tong Wu, Guandao Yang, Zhibing Li, Kai Zhang, Ziwei Liu, Leonidas Guibas, Dahua Lin, and Gordon Wetzstein.
\newblock Gpt-4v(ision) is a human-aligned evaluator for text-to-3d generation.
\newblock In \emph{CVPR}, 2024.

\bibitem[Wu et~al.(2015)Wu, Song, Khosla, Zhang, Tang, and Xiao]{wu2015shape}
Zhirong Wu, Shuran Song, Aditya Khosla, Linguang Zhang, Xiaoou Tang, and Jianxiong Xiao.
\newblock 3d shapenets: A deep representation for volumetric shape modeling.
\newblock In \emph{IEEE Conference on Computer Vision and Pattern Recognition (CVPR)}, Boston, USA, 2015.

\bibitem[Xie et~al.(2024)Xie, Lorraine, Cao, Gao, Lucas, Torralba, Fidler, and Zeng]{xie2024latte3d}
Kevin Xie, Jonathan Lorraine, Tianshi Cao, Jun Gao, James Lucas, Antonio Torralba, Sanja Fidler, and Xiaohui Zeng.
\newblock Latte3d: Large-scale amortized text-to-enhanced3d synthesis.
\newblock \emph{arXiv preprint arXiv:2403.15385}, 2024.

\bibitem[Xu et~al.(2023)Xu, Wang, Cheng, Cao, Shan, Qie, and Gao]{xu2023dream3d}
Jiale Xu, Xintao Wang, Weihao Cheng, Yan-Pei Cao, Ying Shan, Xiaohu Qie, and Shenghua Gao.
\newblock Dream3d: Zero-shot text-to-3d synthesis using 3d shape prior and text-to-image diffusion models.
\newblock In \emph{Proceedings of the IEEE/CVF Conference on Computer Vision and Pattern Recognition}, pages 20908--20918, 2023.

\bibitem[Xu et~al.(2024)Xu, Tan, Luan, Bi, Wang, Li, Shi, Sunkavalli, Wetzstein, Xu, and Zhang]{xu2024dmv3d}
Yinghao Xu, Hao Tan, Fujun Luan, Sai Bi, Peng Wang, Jiahao Li, Zifan Shi, Kalyan Sunkavalli, Gordon Wetzstein, Zexiang Xu, and Kai Zhang.
\newblock Dmv3d: Denoising multi-view diffusion using 3d large reconstruction model.
\newblock In \emph{ICLR}, 2024.

\bibitem[Zhang et~al.(2022)Zhang, Kolkin, Bi, Luan, Xu, Shechtman, and Snavely]{zhang2022arf}
Kai Zhang, Nick Kolkin, Sai Bi, Fujun Luan, Zexiang Xu, Eli Shechtman, and Noah Snavely.
\newblock Arf: Artistic radiance fields.
\newblock In \emph{ECCV}, 2022.

\bibitem[Zhang et~al.(2023)Zhang, Rao, and Agrawala]{zhang2023adding}
Lvmin Zhang, Anyi Rao, and Maneesh Agrawala.
\newblock Adding conditional control to text-to-image diffusion models.
\newblock In \emph{Proceedings of the IEEE/CVF International Conference on Computer Vision}, pages 3836--3847, 2023.

\bibitem[Zhu et~al.(2017)Zhu, Park, Isola, and Efros]{zhu2017unpaired}
{Jun-Yan} Zhu, Taesung Park, Phillip Isola, and Alexei~A Efros.
\newblock Unpaired image-to-image translation using cycle-consistent adversarial networks.
\newblock In \emph{Proceedings of the IEEE international conference on computer vision}, pages 2223--2232, 2017.

\bibitem[Zhu et~al.(2024)Zhu, Zhuang, and Koyejo]{zhu2024hifa}
Junzhe Zhu, Peiye Zhuang, and Sanmi Koyejo.
\newblock {HIFA}: High-fidelity text-to-3d generation with advanced diffusion guidance.
\newblock In \emph{The Twelfth International Conference on Learning Representations}, 2024.

\end{thebibliography}
}
\renewcommand{\thesection}{\Alph{section}}
\setcounter{section}{0} 

\clearpage
\maketitlesupplementary

In this supplementary document, we (1) further discuss our limitations and potential mitigations of the Janus problem using MVDream. We then (2) provide more evaluations of our method and compare it with image-to-3D methods as well as a second style transfer method. We also (3) provide a detailed derivation of our stylized score distillation, and (4) more details of our GPT-based user study. For more qualitative results, please see our supplementary video.

\section{Further discussion on limitations}
\label{sec:limitations_more}

\subsection{Multi-face Janus problem}

Our method inherits the Janus problem~\citep{poole2022dreamfusion} known to score distillation sampling because we use Stable Diffusion, a single-view image diffusion model as the 3D generation prior. To mitigate this problem, one can perform score distillation with multi-view diffusion models. Here, we demonstrate that our stylized score distillation (SSD) extends naturally to MVDream, a popular multi-view diffusion model~\citep{shi2024MVDream}.

\begin{figure}[t]
    \centering
    \includegraphics[width=\linewidth]{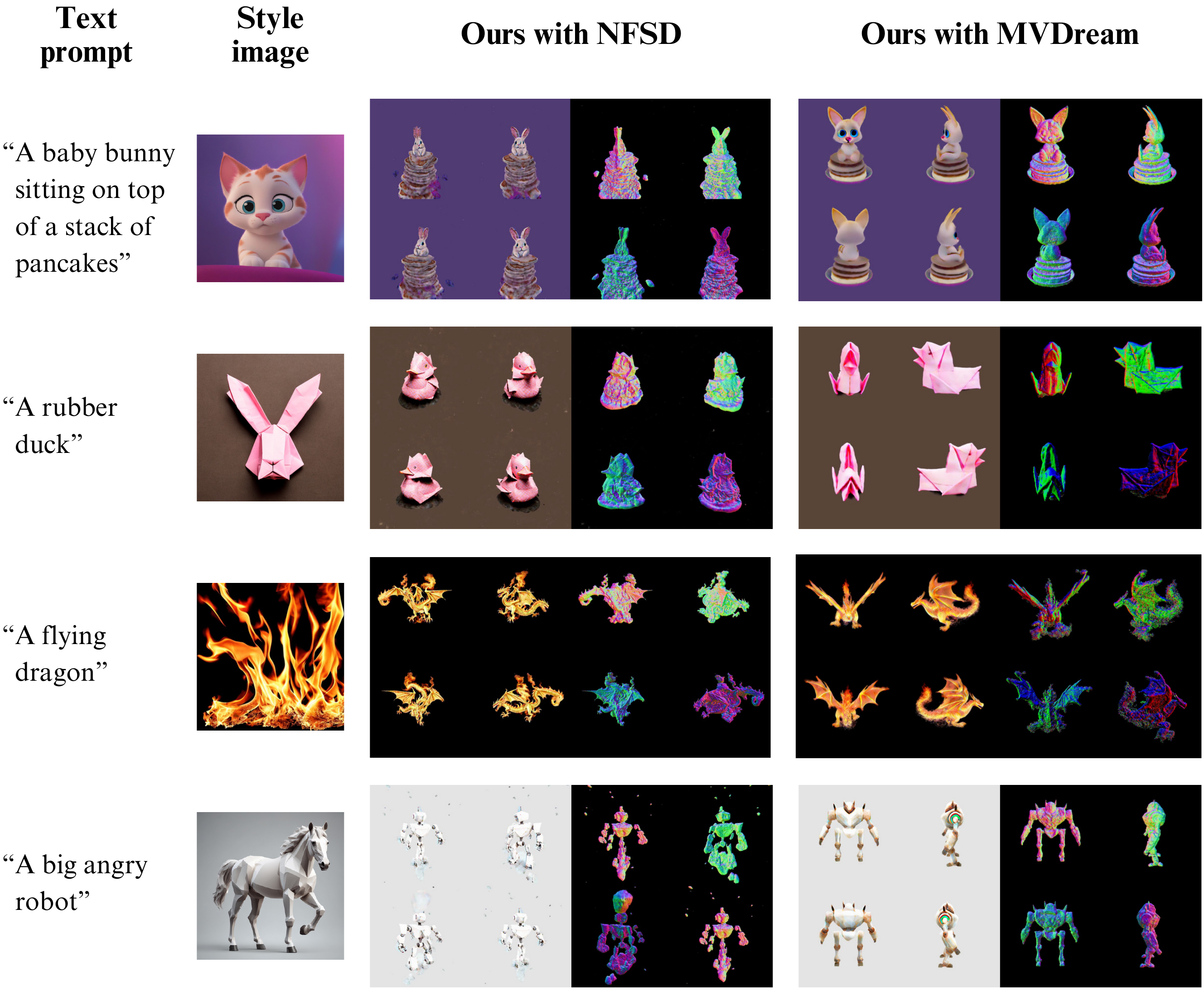}
    \caption{Example cases where multi-face Janus problem occurred and was mitigated by using MVDream.}
    \label{fig:janus}
\end{figure}

In \Cref{fig:janus} we present cases where multi-face Janus problem is mitigated when our method is combined with MVDream. Multi-view videos of these examples can be found in our supplementary video.

\subsection{Discussion on hard cases}
\label{subsec:hard}

We found style images with ambiguous content or complex backgrounds particularly challenging, e.g., fire style from \Cref{fig:janus} or cloud style from \Cref{fig:img-to-3d}. 
We found that our style ratio scheduling is particularly important to address these cases during optimization. 
Without proper scheduling, these cases produced poor results with little or no information about the object in the final results.

\section{More evaluations}

\subsection{Image-to-3D on stylized images}
\begin{figure}[t]
    \centering
    \includegraphics[width=\linewidth]{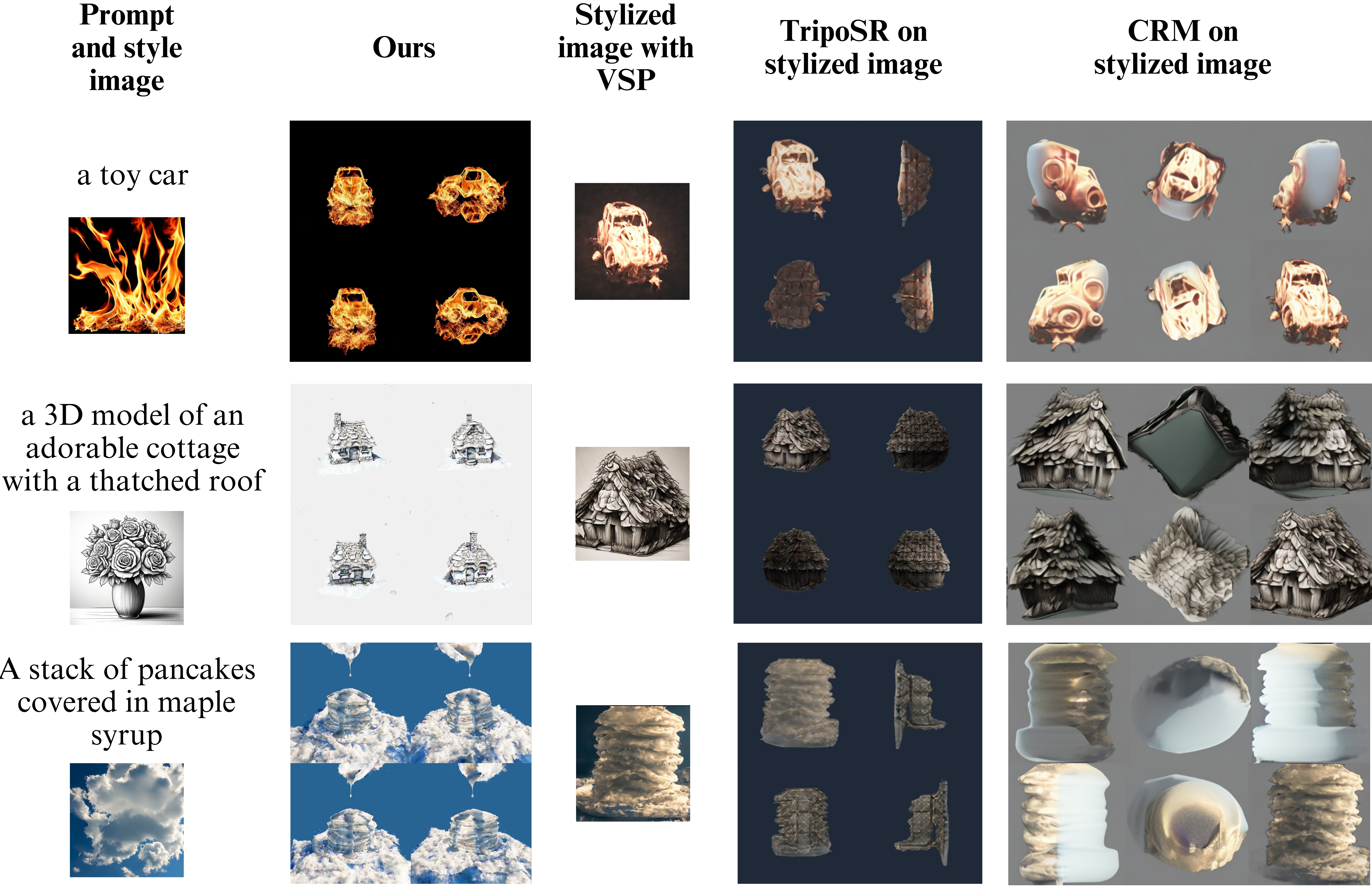}
    \caption{Comparison of our method with results from image-to-3D on stylized images.}
    \label{fig:img-to-3d}
\end{figure}

We experimented with lifting stylized images to 3D using CRM and TripoSR, two popular large reconstruction models for image-to-3D reconstruction. CRM also uses ImageDream~\cite{wang2023imagedream}, an image-conditioned multi-view diffusion model as an initialization to train their reconstruction model. We found that there are two limitations to this approach. First, when the stylized images have ambiguous 2D geometry (fire around a car), the 3D results by CRM and TripoSR are worse than ours. Second, CRM and TripoSR are trained with clean object rendering, which does not generalize well to stylized images with complex visual effects. Comparison of these cases can be found in \Cref{fig:img-to-3d}.

\subsection{Visual Style Prompting vs. StyleAligned}
\begin{figure}[t]
    \centering
    \includegraphics[width=\linewidth]{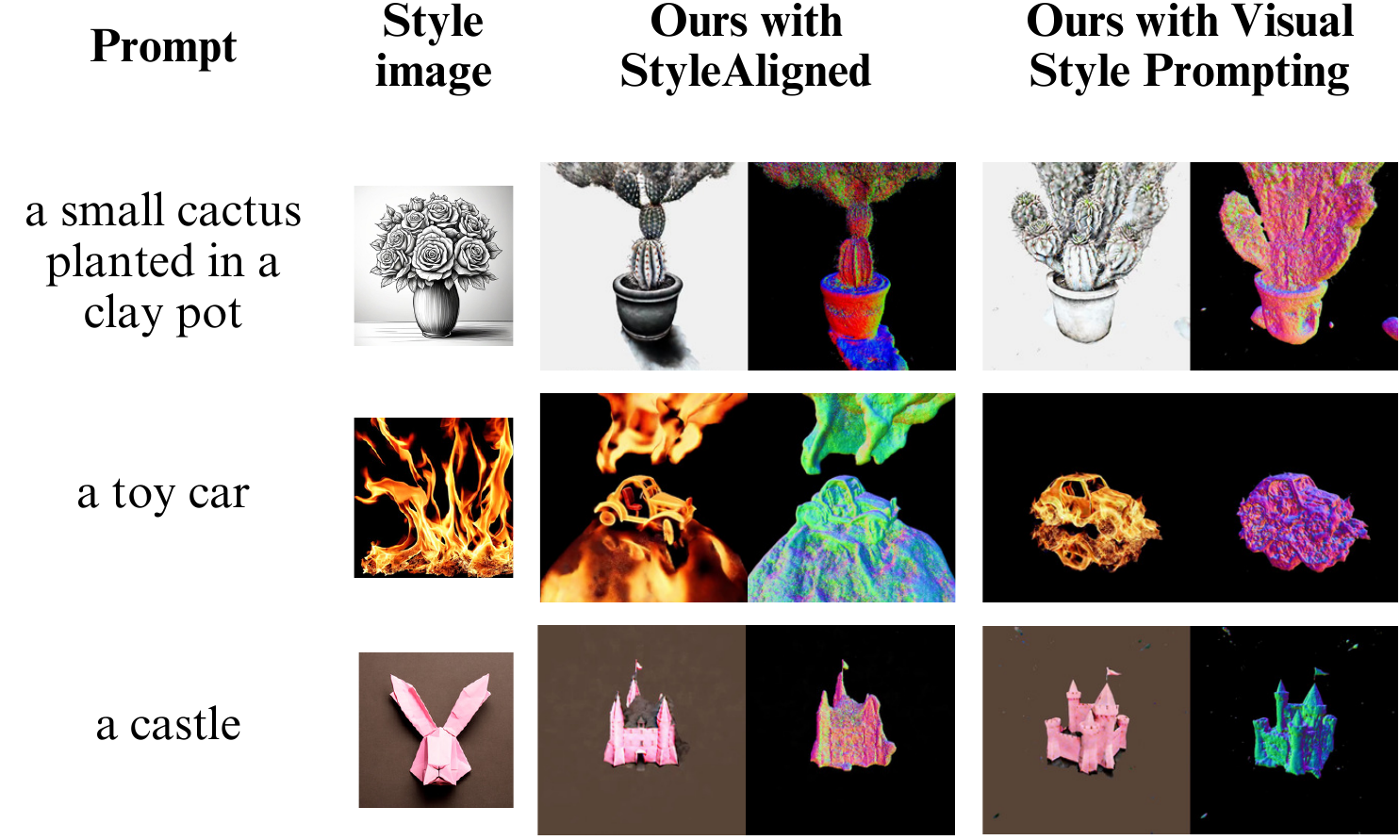}
    \caption{Visual style prompting vs. StyleAligned.}
    \label{fig:style_aligned}
\end{figure}

We additionally tested our score distillation on a different training-free style transfer method using diffusion model. Particularly, we adopt StyleAligned~\cite{hertz2023stylealigned} instead of visual style prompting~\cite{jeong2024visual}. 
We found that our stylized score distillation works well with StyleAligned, and we did not find significant differences in the final results. The biggest difference occurs for harder cases mentioned in~\Cref{subsec:hard} where StyleAligned produced results with more focus on surroundings of generated object. The comparison can be found in~\Cref{fig:style_aligned}.

\newpage
\section{Derivation}
\label{sec:derivation}

\def\s{\mathbf{s}}
\def\z{\mathbf{z}}
\def\x{\mathbf{x}}
\def\I{\mathbf{I}}


Here we provide a detailed derivation of our stylized score distillation. 
Recall that our stylized score distillation aims to minimize $KL(q(\z_t \mid \x = g(\theta)) \parallel p_\phi(\z_t \mid y, \s))$, which is equivalent to
\begin{align}
 \min_{\theta} \mathbb{E}_{\varepsilon} & \left[ \log(q(\z_t \mid \x = g(\theta))) \right. \nonumber \\ & \left. - (1 - \lambda) \log(p_\phi(\z_t \mid y)) - \lambda \log(\hat{p}_\phi(\z_t \mid y, \s)) \right].
\end{align}
Taking the derivative w.r.t. $\theta$ yields a gradient with three terms:
\begin{align}
 \nabla_\theta KL & (q(\z_t \mid \x = g(\theta)) \parallel p_\phi(\z_t \mid y, \s)) \nonumber \\
 & =\mathbb{E}_{\varepsilon} \left[ 
 \underbrace{ \nabla_\theta\log(q(\z_t \mid \x = g(\theta)))}_{(A)} \right. \nonumber \\
 & \left. - (1 - \lambda) \underbrace{\nabla_\theta\log(p_\phi(\z_t \mid y))}_{(B)}
  \right. \nonumber \\
 & \left. - \lambda \underbrace{\nabla_\theta\log(\hat{p}_\phi(\z_t \mid y, \s))}_{(C)}
 \right].
\end{align}
Among these, term (B) and term (C) can be expanded by
\begin{align}
\nabla_\theta\log(p_\phi(\z_t \mid y)) = - \frac{\alpha_t}{\sigma_t} {\varepsilon_{\phi}}(\z_t \mid y) \frac{\partial{\x}}{\partial\theta},
\end{align}
and
\begin{align}
\nabla_\theta\log(\hat{p}_\phi(\z_t \mid y, \s)) = - \frac{\alpha_t}{\sigma_t} \hat{\varepsilon}_{\phi}(\z_t \mid y, \s) \frac{\partial{\x}}{\partial\theta}.
\end{align}

Term (A) can be expanded to
\begin{align}
\nabla_{\theta} \log q(\z_t|\x) &= \left( \underbrace{\frac{\partial \log q(\z_t|\x)}{\partial x}}_{\text{parameter score}} + \underbrace{\frac{\partial \log q(\z_t|\x)}{\partial \z_t} \frac{\partial \z_t}{\partial \x}}_{\text{path derivative}} \right) \alpha_t \frac{\partial \x}{\partial \theta} \nonumber \\
&= \left( \frac{\alpha_t} {\sigma_t} {\epsilon} - \frac{\alpha_t}{ \sigma_t} {\epsilon} \right) \alpha_t \frac{\partial \x}{\partial \theta} = 0.
\end{align}
%
Similarly to DreamFusion~\citep{poole2022dreamfusion}, based on Sticking-the-Landing \citep{roeder2017sticking}, we can discard the parameter score term and keep only the path derivative term. 

The final gradient becomes
\begin{align}
\nabla_{\theta} \mathcal{L}_{SSD} \qquad \qquad \qquad \qquad \qquad \qquad \qquad \qquad \qquad \qquad \nonumber \\
= \mathbb{E}_{t, \z_t\mid \x} \left[ \omega(t) \frac{\sigma_t}{\alpha_t} \nabla_{\theta}KL(q(\z_t \mid \x = g(\theta))
\parallel p_\phi(\z_t \mid y, \s))
 \right] 
\nonumber \\ 
= \mathbb{E}_{t, \varepsilon} \left[ \omega(t) \left( (1 - \lambda) \varepsilon_{\phi}(\z_t \mid y)
+ \lambda \hat{\varepsilon}_{\phi}(\z_t \mid y, \s) - \varepsilon \right) \frac{\partial \x}{\partial \theta} \right].
\end{align}

\newpage
\section{GPT-4 Evaluation Template}
\label{sec:gpt4}

\begin{figure}[t]
    \centering
    \includegraphics[width=\linewidth]{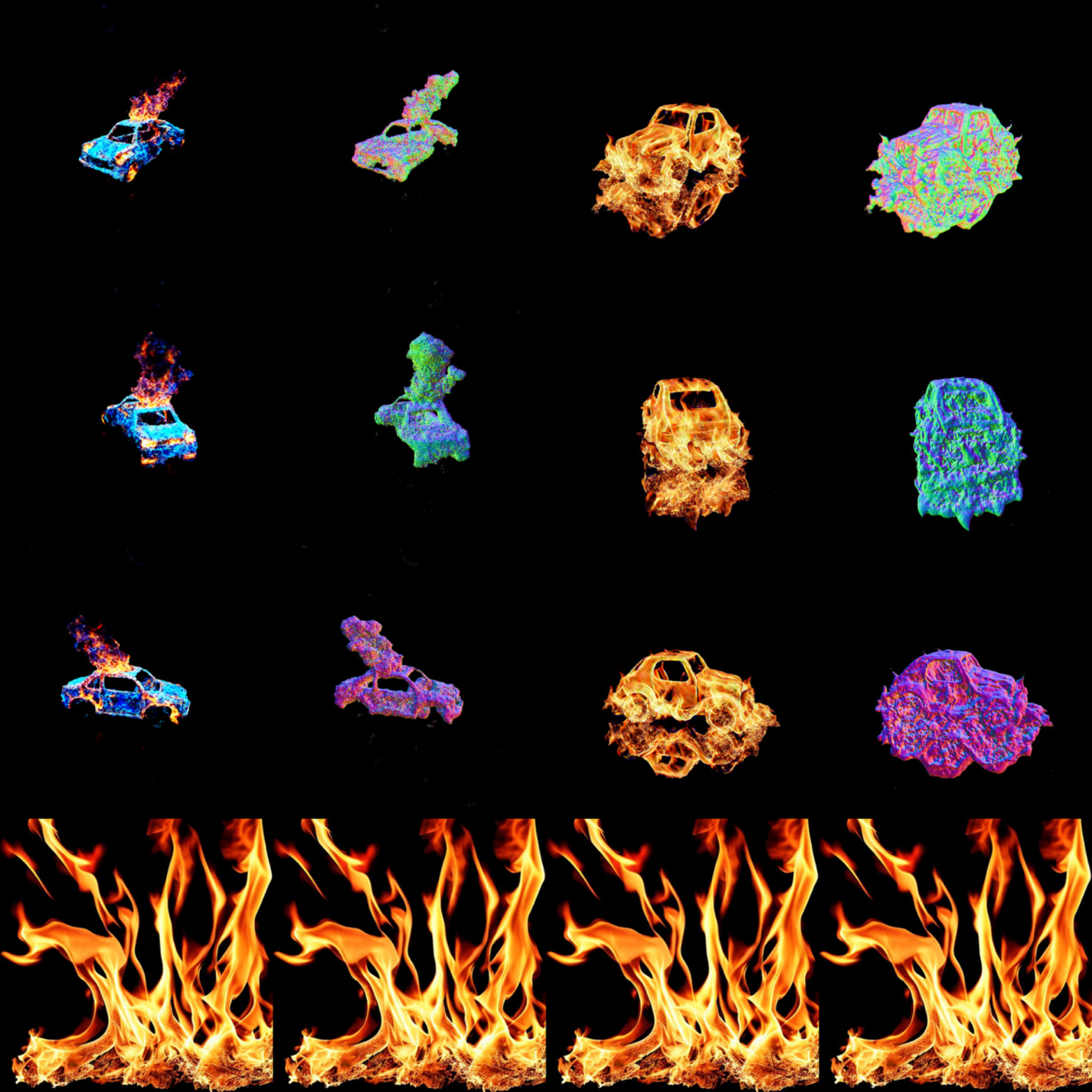}
    \caption{An image grid sent to GPT for evaluation. The first and second column show the color rendering and normal map of the first method, and the third and fourth column are for the second method. Each row shows a camera view of the objects, and the last row shows the style reference. The prompt is "a toy car" and the style image represents fire on a black background. GPT is asked to pick the better result out of the two presented methods.}
    \label{fig:gpt_example}
\end{figure}

We extend the GPTEval3D toolbox~\citep{wu2023gpteval3d} to evaluate the style alignment between the generated 3D objects and the style reference image. 
The toolbox prompts the GPT-4 language model from OpenAI to perform the evaluation.
The entire text prompt to GPT-4 is listed below.
Instruction \#6 is for style alignment evaluation.
\Cref{fig:gpt_example} shows an example of the image grid sent to GPT-4 for evaluation.
In \Cref{table:gpt_results}, we present the detailed Elo scores obtained from this evaluation.

\lstset{breaklines=true}
\begin{lstlisting}
Our task here is the compare two 3D objects, both generated from the same text description.
We want to decide which one is better according to the provided criteria.

I will provide you with a specific multi-view images of two 3D objects, where the left part of it are image renderings and normal renderings of 3D object 1, and the right part denotes those of 3D object 2.
At the bottom of the image, last row, you can see the style image duplicated four times. This image is the reference image for the style of the 3D object.

# Instruction
1. Text prompt and Asset Alignment. Focus on how well they correspond to the given text description. An ideal model should accurately reflect all objects and surroundings mentioned in the text prompt, capturing the corresponding attributes as described. Please first describe each of the two models, and then evaluate how well it covers all the attributes in the original text prompt.

2. 3D Plausibility. Look at both the RGB and normal images and imagine a 3D model from the multi-view images. Determine which model appears more natural, solid, and plausible. Pay attention to any irregularities, such as abnormal body proportions, duplicated parts, or the presence of noisy or meaningless 3D structures. An ideal model should possess accurate proportions, shapes, and structures that closely resemble the real-world object or scene.

3. Geometry-Texture Alignment. This examines how well the texture adheres to the geometry. The texture and shape should align with each other locally. For instance, a flower should resemble a flower in both the RGB and normal map, rather than solely in the RGB. The RGB image and its corresponding normal image should exhibit matching structures.

4. Low-Level Texture Details. Focus on local parts of the RGB images. Assess which model effectively captures fine details without appearing blurry and which one aligns with the desired aesthetic of the 3D model. Note that overly abstract and stylized textures are not desired unless specifically mentioned in the text prompt.

5. Low-Level Geometry Details. Focus on the local parts of the normal maps. The geometry should accurately represent the intended shape. Note that meaningless noise is not considered as high-frequency details. Determine which one has a more well-organized and efficient structure, which one exhibits intricate details, and which one is more visually pleasing and smooth.

6. Style Image Alignment. Look at the style image at the bottom and determine which model better aligns with the desired style. Do you see any patterns from style image that are present in any of 3D objects? 3D object should ideally represent the provided prompt, but in the style from the style image. It should be a good combination of the prompt and reference style.

7. Considering all the degrees above, which one is better overall?

Take a really close look at each of the multi-view images for these two 3D objects before providing your answer.

When evaluating these aspects, focus on one of them at a time.

Try to make independent decisions between these criteria.

# Output format
To provide an answer, please provide a short analysis for each of the abovementioned evaluation criteria.
The analysis should be very concise and accurate.

For each of the criteria, you need to make a decision using these three options:
1. Left (object 1) is better;
2. Right (object 2) is better;
3. Cannot decide.
IMPORTANT: PLEASE USE THE THIRD OPTION SPARSELY.

Then, in the last row, summarize your final decision by "<option for criterion 1> <option for criterion 2> <option for criterion 3> <option for criterion 4> <option for criterion 5> <option for criterion 6> <option for criterion 7>".

An example output looks like follows:
"
Analysis:
1. Text prompt & Asset Alignment: The left one xxxx; The right one xxxx;
The left/right one is better or cannot decide.

2. 3D Plausibility. The left one xxxx; The right one xxxx;
The left/right one is better or cannot decide.

3. Geometry-Texture Alignment. The left one xxxx; The right one xxxx;
The left/right one is better or cannot decide.

4. Low-Level Texture Details. The left one xxxx; The right one xxxx;
The left/right one is better or cannot decide.

5. Low-Level Geometry Details. The left one xxxx; The right one xxxx;
The left/right one is better or cannot decide.

6. Style Image Alignment. The left one xxxx; The right one xxxx;
The left/right one is better or cannot decide.

7. Overall, xxxxxx
The left/right one is better or cannot decide.


Final answer:
x x x x x x x (e.g., 1 2 2 3 2 1 1/ 3 3 3 2 1 3 3 / 3 2 2 1 1 1 1)
"

Following is the text prompt from which these two 3D objects are generated:
"<PROMPT>"
Please compare these two 3D objects as instructed.

\end{lstlisting}

\begin{table*}
\centering 
\footnotesize
\begin{tabular}{lccccccc}

\toprule


\textbf{Methods} & \vtop{\hbox{\strut \textbf{Text-Asset}}\hbox{\strut \textbf{Alignment}}} & \textbf{3D Plausibility} & \vtop{\hbox{\strut \textbf{Text-Geometry}}\hbox{\strut \textbf{Alignment}}} & \textbf{Texture Details} & \textbf{Geometry Details} & \textbf{Style Alignment} & \textbf{Overall} \\
\midrule
Style-in-prompt & 1000.000 & 1000.000 & 1000.000 & 1000.000 & 1000.000 & 1000.000 & 1000.000 \\
Neural style loss & 1022.913 & 1045.297 & 1063.513 & 1039.004 & 1058.329 & 960.737 & 1038.849 \\
Textual inversion & 1035.892 & 1037.247 & 1045.499 & 1028.978 & 1039.247 & 961.984 & 1025.917 \\
\midrule
Ours & \textbf{1118.967} & \textbf{1161.566} & \textbf{1158.723} & \textbf{1150.614} & \textbf{1162.029} & \textbf{1046.029} & \textbf{1140.604} \\
\bottomrule
\end{tabular}
\label{table:gpt_results}
\caption{Detailed results from GPTEval3D evaluation.}
\end{table*}

\end{document}